\documentclass[journal, compsoc]{IEEEtran}

\usepackage{cite}
\usepackage{amsmath,amssymb,amsfonts}
\usepackage{algorithmic}
\usepackage{array}
\usepackage{graphicx}
\usepackage{textcomp}
\usepackage[table]{xcolor}
\usepackage{xcolor}
\usepackage{orcidlink}
\usepackage{multirow}
\usepackage[inline]{enumitem}
\usepackage{caption}
\usepackage{booktabs}
\usepackage{pifont}
\usepackage{arydshln}
\usepackage{multicol}

\def\BibTeX{{\rm B\kern-.05em{\sc i\kern-.025em b}\kern-.08em
    T\kern-.1667em\lower.7ex\hbox{E}\kern-.125emX}}
    
\usepackage{listings}

\newcommand{\cmark}{\ding{51}}  
\newcommand{\xmark}{\ding{55}}  

\definecolor{codegreen}{rgb}{0,0.6,0}
\definecolor{codegray}{rgb}{0.5,0.5,0.5}
\definecolor{codepurple}{rgb}{0.58,0,0.82}
\definecolor{backcolour}{rgb}{1,1,1}

\lstdefinestyle{mystyle}{
    backgroundcolor=\color{backcolour},   
    commentstyle=\color{codegreen},
    keywordstyle = {\color{blue}},
    keywordstyle = [2]{\color{cyan}},
    keywordstyle = [3]{\color{yellow}},
    keywordstyle = [4]{\color{teal}},
    numberstyle=\tiny\color{codegray},
    stringstyle=\color{codepurple},
    basicstyle=\ttfamily\footnotesize,
    breakatwhitespace=false,         
    breaklines=true,                 
    captionpos=b,                    
    keepspaces=true,                 
    numbers=none,                    
    numbersep=5pt,                  
    showspaces=false,                
    showstringspaces=false,
    showtabs=false,                  
    tabsize=2,
    frame=single
}

\title{FaceLiVTv2: An Improved Hybrid Architecture for Efficient Mobile Face Recognition \\
}
\author{Novendra Setyawan, \IEEEmembership{Student Member, IEEE}, Chi-Chia Sun, \IEEEmembership{Member, IEEE}, \\ Mao-Hsiu Hsu, \IEEEmembership{Member, IEEE}, Wen-Kai Kuo,  \IEEEmembership{Member, IEEE}, Jun-Wei Hsieh \IEEEmembership{Senior Member, IEEE} \\
\IEEEauthorblockA{} 
\thanks{Novendra Setyawan, Mao-Hsiu Hsu and Wen-Kai Kuo are with Department of Electro-Optics, National Formosa University, Taiwan;

Novendra Setyawan also with Department of Electrical Engineering University of Muhammadiyah Malang, Indonesia; 

Chi-Chia Sun is with Department of Electrical Engineering, National Taipei University, Taiwan; 

Jun-Wei Hsieh is with College of Artificial Intelligence and Green Energy, National Yang Ming Chiao Tung University, Taiwan; 

Corresponding Author is Chi-Chia Sun (\textit{E-mail: chichiasun@gm.ntpu.edu.tw})}}

\begin{document}
\maketitle
\markboth{IEEE Transactions on Biometrics, Behavior, and Identity Science}%
{Setyawan \etal{}: FaceLiVTv2}

\begin{abstract}

Lightweight face recognition is increasingly important for deployment on edge and mobile devices, where strict constraints on latency, memory, and energy consumption must be met alongside reliable accuracy. Although recent hybrid CNN-Transformer architectures have advanced global context modeling, striking an effective balance between recognition performance and computational efficiency remains an open challenge. In this work, we present FaceLiVTv2, an improved version of our FaceLiVT hybrid architecture designed for efficient global--local feature interaction in mobile face recognition. At its core is Lite MHLA, a lightweight global token interaction module that replaces the original multi-layer attention design with multi-head linear token projections and affine rescale transformations, reducing redundancy while preserving representational diversity across heads. We further integrate Lite MHLA into a unified RepMix block that coordinates local and global feature interactions and adopts global depthwise convolution for adaptive spatial aggregation in the embedding stage. Under our experimental setup, results on LFW, CA-LFW, CP-LFW, CFP-FP, AgeDB-30, and IJB show that FaceLiVTv2 consistently improves the accuracy-efficiency trade-off over existing lightweight methods. Notably, FaceLiVTv2 reduces mobile inference latency by 22\% relative to FaceLiVTv1, achieves speedups of up to 30.8\% over GhostFaceNets on mobile devices, and delivers 20-41\% latency improvements over EdgeFace and KANFace across platforms while maintaining higher recognition accuracy. These results demonstrate that FaceLiVTv2 offers a practical and deployable solution for real-time face recognition. Code is available at \href{https://github.com/novendrastywn/FaceLiVT}{\textit{https://github.com/novendrastywn/FaceLiVT}}.
\end{abstract}
\section{Introduction}
Face recognition (FR) has become one of the most widely adopted biometric technologies, powering a variety of applications ranging from smartphone authentication and financial transactions to public security and access control \cite{Gururaj2024ReviewFace, Feng2022Detect}. In mobile and embedded scenarios, face verification is often performed locally without relying on cloud services, which imposes strict requirements on computational efficiency, memory footprint, and energy consumption. Thus, designing accurate yet lightweight FR models has emerged as a key challenge in the field of computer vision.

\begin{figure}[t!]
    \centering
    \includegraphics[width=\columnwidth]{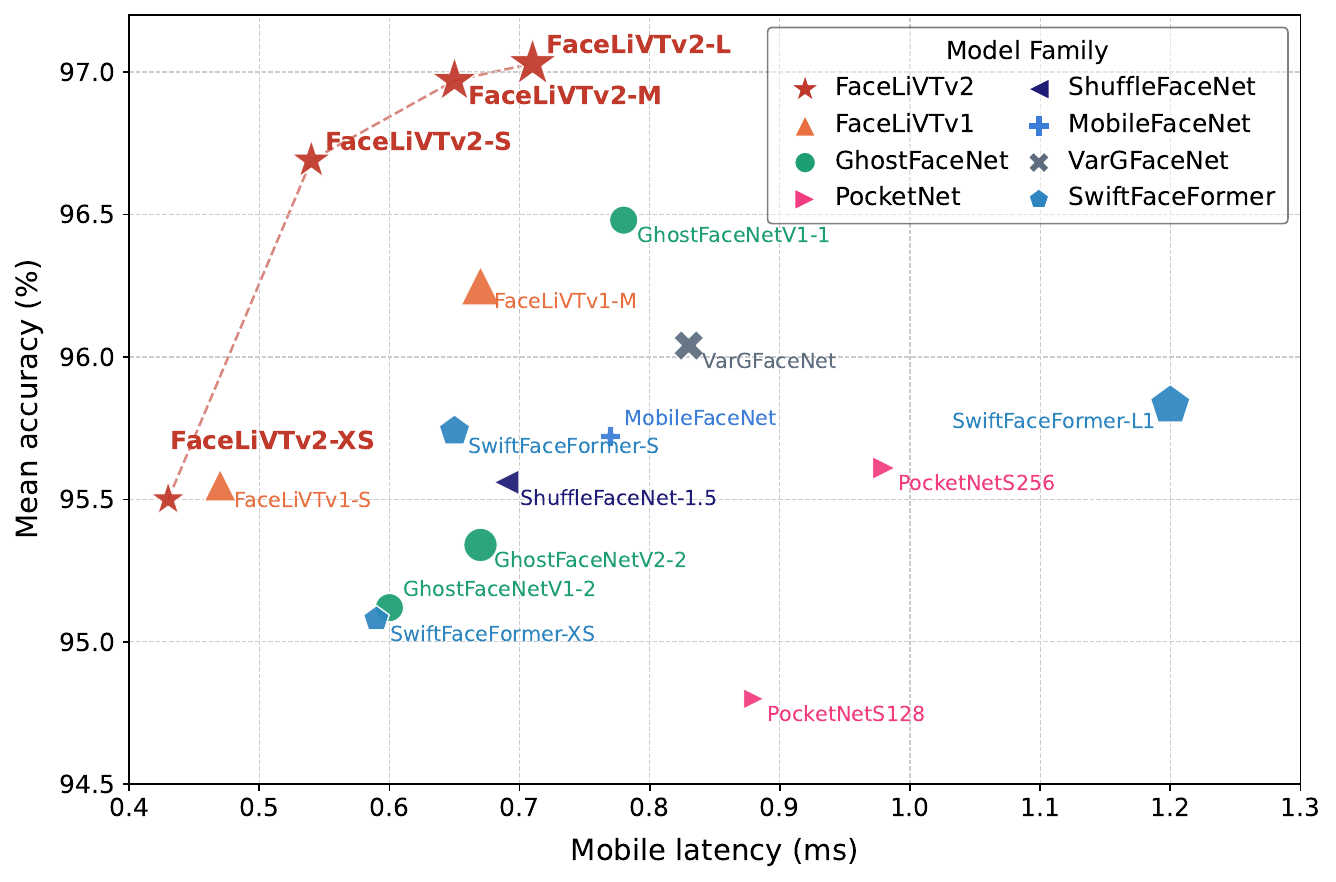}
    \caption{Comparison of our proposed FaceLiVTv2 with other lightweight FR models, including EdgeFace\cite{george2024edgeface}, SwiftFaceFormer\cite{luevano2024swiftfaceformer}, and KANFace\cite{pham2025kanface}. Mean accuracy of Face Benchmark v.s. Mobile Latency (iPhone 15 Pro). The upper-left region of the plot represents the most desirable operating point, where high accuracy is achieved with minimal latency.}
    \label{fig:acc_lat}
\end{figure} 

\begin{table*}[t]
\centering
\caption{Key architectural properties of representative lightweight face recognition models}
\label{tab:arch_property}
\setlength{\tabcolsep}{2.5pt}
\begin{tabular}{m{2.6cm}ccccccm{9.8cm}}
\hline
\multirow{2}{*}{\textbf{Methods}} & \multicolumn{2}{c}{\textbf{Context}} & \textbf{Low} & \textbf{High} & \multicolumn{2}{c}{\textbf{Fast Inference}} & \multicolumn{1}{c}{\multirow{2}{*}{\textbf{Approach}}}\\
\cline{2-3} \cline{6-7}
& \textbf{Global} & \textbf{Local} & \textbf{Comp.} & \textbf{Acc.} & \textbf{Edge} & \textbf{Mobile} & \\ 
\hline
MobileFaceNet~\cite{chen2018mobilefacenets}     & \xmark & \cmark & \cmark & \xmark & \cmark & \cmark & 
Lightweight CNN optimized for mobile face recognition using depthwise separable convolutions \cite{howard2017mobilenets}. \\
GhostFaceNet~\cite{alansari2023ghostfacenets}   & \xmark & \cmark & \cmark & \xmark & \cmark & \cmark & 
CNN-based face recognition model leveraging ghost modules \cite{han2020ghostnet} to reduce redundancy. \\
TransFace~\cite{dan2023transface}               & \cmark & \xmark & \xmark & \cmark & \xmark & \xmark & 
Vanilla Vision Transformer\cite{dosovitskiy2020image} for face recognition with patch-level data augmentation.\\
EdgeFace~\cite{george2024edgeface}               & \cmark & \cmark & \cmark & \cmark & \cmark & \xmark & EdgeNext\cite{maaz2022edgenext}-based architecture with low-rank linear layer. \\
KANFace~\cite{pham2025kanface} & \cmark & \cmark & \cmark & \cmark & \cmark & \xmark & 
Hybrid EdgeFace-based \cite{george2024edgeface} architecture incorporating Kolmogorov–Arnold Networks \cite{liu2024kan} for enhanced representation. \\
SwiftFaceFormer~\cite{luevano2024swiftfaceformer} & \cmark & \cmark & \cmark & \xmark & \cmark & \cmark & 
SwiftFormer\cite{shaker2023swiftformer}-based, Hybrid CNN–Transformer model with additive attention and knowledge distillation. \\
FaceLiVTv1~\cite{Setyawan2025FaceLiVT, setyawan2025facelivtapccas} & \cmark & \cmark & \cmark & \xmark & \cmark & \cmark &
Hybrid CNN-Transformer with reparameterized local token mixing and Multi Head Linear Attention (MHLA), the multi MLP-style token interaction \\
\rowcolor{gray!12}
\textbf{FaceLiVTv2 (Ours)}                & \cmark & \cmark & \cmark & \cmark & \cmark & \cmark &
Co-design architecture based on FaceLiVTv1 \cite{Setyawan2025FaceLiVT, setyawan2025facelivtapccas} by integrating reparameterized local token mixing with a lightweight MHLA (Lite MHLA) and adaptive spatial aggregation head for efficient joint global–local feature learning. \\
\hline
\end{tabular}
\end{table*}

In the last decade, deep convolutional neural networks (CNNs) have significantly advanced FR performance, achieving near-human accuracy on unconstrained benchmarks. However, this progress has come at the expense of increasingly complex architectures with tens of millions of parameters and billions of floating-point operations (FLOPs), such as ResNet50-ArcFace\cite{deng2019arcface, zhang2023texture}, making direct deployment on edge devices impractical. To address this, researchers have developed compact CNN-based architectures optimized for mobile deployment. Notable examples include MobileFaceNet\cite{chen2018mobilefacenets}, ShuffleFaceNet\cite{martinez2021benchmarking}, VarGFaceNet\cite{yan2019vargfacenet}, MixFaceNet\cite{boutros2021mixfacenets}, and GhostFaceNet\cite{alansari2023ghostfacenets}, each of which adopts different design strategies to reduce 
parameters as well as computational complexity while maintaining high levels of accuracy. These lightweight CNNs have demonstrated impressive speed-accuracy trade-offs, but their limited receptive fields often hinder their ability to capture long-range facial dependencies.

In parallel, Vision Transformers (ViTs)\cite{dosovitskiy2020image} have emerged as powerful alternatives for computer vision tasks, offering global receptive fields and superior capabilities in modeling long-range relationships. Early explorations such as Face Transformer \cite{zhong2021face} and TransFace\cite{dan2023transface} showed that transformer-based FR models could rival state-of-the-art CNNs in accuracy. However, their computational complexity regarding the attention mechanism makes them prohibitively expensive for mobile applications, with parameter counts exceeding hundreds of millions and latencies far beyond real-time constraints. To mitigate these drawbacks, several hybrid CNN–Transformer architectures have been introduced. EdgeFace\cite{george2024edgeface} and SwiftFaceFormer\cite{luevano2024swiftfaceformer}, for example, adapt the EdgeNeXt\cite{maaz2022edgenext} and SwiftFormer\cite{shaker2023swiftformer} backbones, striking a balance between contextual modeling and linear complexity. KANFace \cite{pham2025kanface} proposed a hybrid model with Kolmogorov-Arnold Networks to balance model capacity and feature discriminability. These works highlight the potential of hybrid designs, but they often face trade-offs between model compactness, discriminative ability, and inference efficiency.

Recently, FaceLiVT~\cite{Setyawan2025FaceLiVT, setyawan2025facelivtapccas} introduced a hybrid CNN-Transformer architecture that combines structural reparameterization with Multi-Head Linear Attention (MHLA) to reduce computational complexity while maintaining competitive accuracy. However, several aspects of its design leave room for improvement. The first-generation MHLA adopts an MLP-style two-layer linear design with a non-linear activation, which introduces additional overhead. The integration of local and global token interactions remains suboptimal, and the embedding head relies on uniform global pooling, which limits adaptive spatial aggregation. These observations motivate a re-examination of global token interaction, local-global feature
integration, and embedding design.

In this paper, we revisit FaceLiVT~\cite{Setyawan2025FaceLiVT, setyawan2025facelivtapccas} (hereafter referred to as FaceLiVTv1) and propose FaceLiVTv2, a more accurate and efficient architecture tailored for real-time FR on mobile and embedded platforms. FaceLiVTv2 co-designs three complementary components: reparameterized local token mixing (RepMix) to strengthen local-global feature interaction while preserving efficient inference; a lightweight multi-head linear global modeling module to capture long-range dependencies with lower complexity than its predecessor; and an adaptive spatial aggregation head to replace uniform global pooling. At the core of these improvements, we have redesigned the global modeling module as Lite MHLA, a streamlined multi-head token interaction formulated as a single linear mapping with an affine rescaling transformation that reduces parameter redundancy and stabilizes optimization while preserving representational diversity across heads. All architectural choices are guided by the efficiency constraints of edge and mobile face recognition.

The main contributions of this work are as follows:
\begin{enumerate}
    \item We revisit the MHLA module and propose Lite MHLA, an optimized
    formulation that reduces redundancy while improving global context
    modeling over the original FaceLiVT design.
    \item We present FaceLiVTv2, a co-designed architecture integrating a
    refined embedding head, reparameterized convolutional mixers, and the
    Lite MHLA module, yielding a better balance between accuracy and
    inference latency.
    \item We conducted extensive experiments on multiple challenging FR
    benchmarks (LFW, CA-LFW, CP-LFW, CFP-FP, AgeDB-30, IJB-B, IJB-C, TinyFace),
    demonstrating that FaceLiVTv2 outperforms both its predecessor and
    recent lightweight state-of-the-art models.
    \item We validate that our models maintain edge and mobile-friendliness
    by profiling latency on widely used edge and mobile devices, as depicted in Fig~\ref{fig:acc_lat}.
\end{enumerate}

\begin{figure*}[!ht]
    \centering
    \includegraphics[width=14cm]{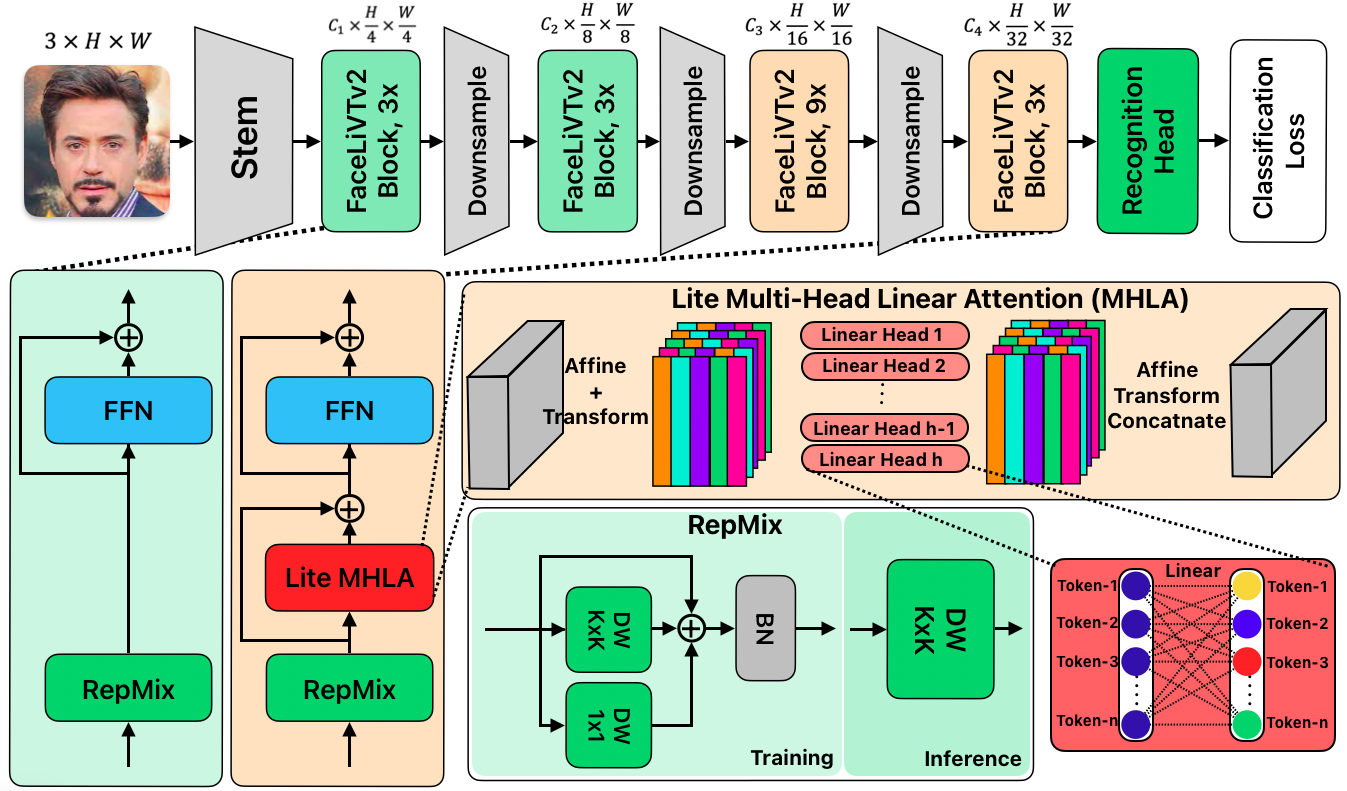}
    \caption{FaceLiVTv2 architecture with Lite MHLA and structural reparameterization. Stages 1 and 2 use RepMix
    and the last stage combines RepMix with Lite MHLA as spatial mixer.}
    \label{fig:FaceLiVTv2}
\end{figure*}
 This paper is organized as follows. Section~II reviews related work on lightweight CNNs, transformers, and hybrid based FR. Section~III presents the proposed FaceLiVTv2 architecture in detail. Section~IV describes the experimental setup, benchmark datasets, evaluation results, and ablation studies. Section~V concludes the paper
and outlines future directions. For clarity, the following abbreviations are used throughout this paper:

\begin{table}[h]
\small
\begin{tabular}{ll}
\textbf{CNN}: & Convolutional Neural Network. \\
\textbf{FR}: & Face Recognition \\
\textbf{FFN}: & Feed Forward Network \\
\textbf{GDConv}: & Global Depthwise Convolution \\
\textbf{MHSA}: & Multi-Head Self-Attention \\
\textbf{MHLA}: & Multi-Head Linear Attention \\
\textbf{MLP}: & Multi Layer Perceptron \\
\textbf{RepMix}: & Reparameterize Token Mixer \\
\textbf{SOTA}: & State-of-The-Art \\
\end{tabular}
\end{table}

\section{Related Works}
Over the past decade, deep learning–based FR has achieved remarkable progress, driven by advances in network architectures, large-scale datasets, and discriminative loss functions such as ArcFace \cite{deng2019arcface}, CosFace \cite{wang2018cosface}, MagFace \cite{meng2021magface}, ElasticFace \cite{boutros2022elasticface}, AdaFace\cite{kim2022adaface}, and TopoFR\cite{dan2024topofr}. Beyond loss design, recent studies have explored complementary strategies to improve FR robustness in unconstrained settings. ARFace~\cite{Zhang2022ARFace} introduces reinforcement learning-based attention regularization that dynamically focuses on informative facial
regions, while CATFace~\cite{Alipour2024CATFace} proposes a cross-attribute-guided transformer with self-attention distillation, leveraging soft biometric attributes to enhance recognition in low-quality and surveillance scenarios. Addressing degraded inputs from a different angle, Zhang et al.~\cite{zhang2023texture} employ texture-guided transfer learning to bridge the domain gap between high- and low-quality face images, while Khalid et al.~\cite{khalid2020resolution} tackle resolution mismatch through a distillation framework that learns resolution-invariant embeddings. However, as FR systems move from cloud deployment to mobile and embedded platforms, researchers have increasingly focused on designing lightweight models that provide a favorable trade-off between accuracy and efficiency. In this section, we review related efforts in three directions: lightweight CNN, Vision Transformer–based FR, and hybrid architectures, along with attribute-guided and regularization approaches that improve robustness in unconstrained scenarios.

\subsection{Lightweight CNNs for Face Recognition}
The development of lightweight CNN architectures has been a major strategy for enabling real-time face recognition (FR) on mobile and embedded platforms. These models are typically derived from efficient convolutional backbones originally designed for general-purpose vision tasks, with modifications to improve their discriminative power for FR.
MobileFaceNet \cite{chen2018mobilefacenets} adapts MobileNetV2 with architectural adjustments to achieve mobile level FR, significantly reducing model size while achieving competitive performance on the Labeled Faces in the Wild (LFW) \cite{huang2008labeled} and AgeDB-30\cite{moschoglou2017agedb} benchmarks. ShuffleFaceNet \cite{martindez2019shufflefacenet} extends ShuffleNetV2 with channel shuffling and efficient downsampling to reduce computational costs. VarGFaceNet \cite{yan2019vargfacenet} leverages variable group convolutions to enhance feature diversity while maintaining compactness, and MixFaceNets \cite{boutros2021mixfacenets} incorporate mixed depthwise kernels and channel shuffle operations for improved efficiency. More recently, GhostFaceNets \cite{alansari2023ghostfacenets} exploit inexpensive operations to generate redundant feature maps, achieving strong performance with fewer FLOPs.
These lightweight CNNs consistently provide accurate results on standard benchmarks, often with parameter counts in the range of 0.5M–7M. However, their limited receptive fields and inability to model long-range dependencies restrict their effectiveness in challenging scenarios, such as large pose variations, occlusions, or poor illumination.

\begin{figure*}
    \centering
    \includegraphics[width=18cm]{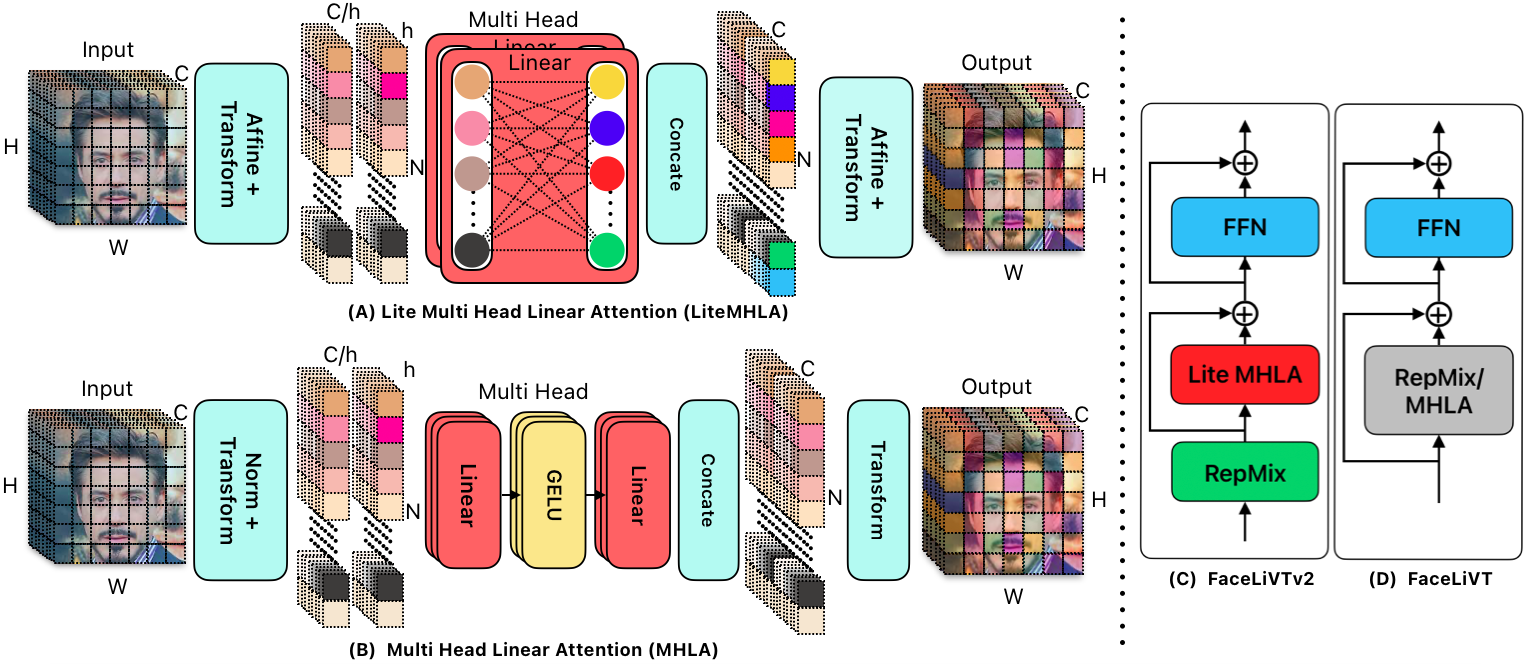}
    \caption{Architecture comparison of (a) LiteMHLA in FaceLiVTv2 and (b) MHLA in FaceLiVTv1, (c) FaceLiVTv2 Block and (d) FaceLiVT Block. C, H and W denotes as the number of channel, height and width of feature map, h denotes as the number of head on the MHLA.}
    \label{fig:v1_vs_v2}
\end{figure*} 

\subsection{Lightweight Transformers for Face Recognition}
Transformers have recently been applied to FR due to their ability to capture long-range dependencies through global self-attention. However, the higher complexity of standard MHSA poses challenges for resource-limited deployment. To address this, several lightweight or efficient Transformer-based FR approaches have been proposed.

Face Transformer \cite{zhong2021face} was the first to apply ViT to FR, introducing a sliding-patch design to better capture inter-patch relationships. TransFace \cite{dan2023transface, dan2025transface++} calibrated transformer training from a data-centric perspective, improving stability and generalization across large-scale benchmarks. Other variants have integrated efficient attention mechanisms or token reduction strategies to reduce computational overhead. For instance, MobileFaceFormer \cite{li2024mobilefaceformer} combines CNN and Transformer branches with convolutional token initialization for more efficient local-global feature extraction, while EdgeFace \cite{george2024edgeface} incorporates a low-rank linear (LoRaLin) module to approximate attention and reduce costs. SwiftFaceFormer \cite{luevano2024swiftfaceformer} further explored additive attention with linear complexity, demonstrating competitive accuracy with reduced latency. KANsFace~\cite{pham2025kanface} integrated Kolmogorov-Arnold Networks (KANs) into face recognition architectures by replacing the conventional MLP-based head with a KANs module in the EdgeFace-based model. Overall, lightweight ViTs and hybrid CNN–Transformer models demonstrate that efficient global context modeling is possible, but existing designs still face trade-offs between discriminative capacity and strict resource constraints. This motivates further refinement of lightweight attention mechanisms, such as the MHLA adopted in our FaceLiVTv1 framework \cite{Setyawan2025FaceLiVT, setyawan2025facelivtapccas}, and its improved variant in FaceLiVTv2. As summarized in Table \ref{tab:arch_property}, the extracted properties from effective face recognition architectures serve as design principles for constructing models to achieve a favorable trade-off between accuracy, efficiency, and scalability in cross-platform and low-computation scenarios.

\section{Proposed Method}
FaceLiVTv1 demonstrated competitive performance among lightweight Transformer-based face recognition models. However, several limitations affected its scalability and robustness in unconstrained scenarios. First, although computationally efficient, the original MHLA formulation introduced redundant linear token projections across heads and provided limited cross-token interaction, which constrained its ability to model fine-grained dependencies arising from complex variations in pose, illumination, and occlusion. Second, the embedding head relied on global average pooling, resulting in uniform spatial aggregation that weakened discriminative local cues important for open-set face recognition. Third, the separate branch design of RepMix blocks limited their integration with global token interaction, reducing overall architectural coherence and efficiency.
To address these limitations, we propose FaceLiVTv2, an improved variant that incorporates three key architectural refinements: (1) a redesigned MHLA module (called \textit{Lite MHLA}) with simplified linear token interaction and affine feature rescale transformation  for more efficient global dependency modeling, (2) a \textit{GDConv} embedding head that enables adaptive spatial aggregation for more discriminative feature representations, and (3) a \textit{unified RepMix–Lite MHLA structure} that strengthens local–global feature integration while maintaining inference efficiency.

\subsection{Architecture}
The next iteration of the FaceLiVT macro-design continues to adopt a feature pyramid architecture~\cite{liu2023efficientvit,pan2022edgevits} to balance multi-scale representation and efficiency. Unlike FaceLiVTv1, which applies the same spatial and channel mixing modules across all stages, FaceLiVTv2 introduces a stage-specific block design, as illustrated in Fig.~\ref{fig:FaceLiVTv2}. Specifically, the first two stages employ a combination of RepMix and FFN blocks, while the final two stages adopt three stacked residual blocks consisting of RepMix, Lite MHLA, and FFN.

This design reflects the observation that early-stage features mainly capture low-level spatial patterns, where global token interaction offers limited benefits and introduces unnecessary overhead. In contrast, higher-level features benefit more from global modeling; therefore, RepMix is followed by Lite MHLA in the later stages to jointly capture local structural information and long-range token interactions.

More precisely, let $X_i \in \mathbb{R}^{H_i \times W_i \times C_i}$ denote the feature map at stage $i$ with spatial resolution $H_i \times W_i$ and $C_i$ channels. The detailed formulation of each block and its associated operators is provided in Eq.~(\ref{eq:meta}).
\begin{equation}
    \begin{split}
    X_i &= RepMix(X_i), \\
    X_i &= X_i + LiteMHLA(X_i), \\
    X_i &= X_i + FFN(X_i).
    \end{split}
\label{eq:meta}
\end{equation}
The FFN layer is composed of two linearly point-wise convolution layers and a single activation function that can be expressed in Eq. (\ref{eq:ffn}) as follows:
\begin{equation}
    FFN(X_i)=BN\Big(\sigma\big(BN(X_i*W_e)\big)*W_r\Big),
    \label{eq:ffn}
\end{equation}
where $W_e\in\mathbb{R}^{C_i\times rC_i \times 1 \times 1}$ and $W_r\in\mathbb{R}^{(rC_i)\times C_i \times 1 \times 1}$ are the weights  of the FFN layer, with a default expansion ratio value of $r=2$. $GELU(.)$ is chosen as the activation function ($\sigma$) in the middle of the FFN. 

\subsection{Global Depth-wise Convolution (GDConv) Head}
Although global token interaction is performed prior to aggregation, the uniform Global Average Pooling (GAP) commonly used as the embedding head in Transformer-based~\cite{dan2023transface,dan2025transface++} and hybrid CNN–Transformer face recognition models~\cite{george2024edgeface,maaz2022edgenext,shaker2023swiftformer}, including FaceLiVT~\cite{Setyawan2025FaceLiVT}, treats all spatial locations equally and overlooks channel-specific spatial relevance. Such uniform aggregation may suppress discriminative local cues that are critical for open-set face recognition. In FaceLiVTv2, we refine the embedding head by replacing GAP with a GDConv layer, enabling adaptive spatial feature aggregation. GDConv complements global token interaction by learning channel-wise spatial weighting, allowing the model to emphasize discriminative regions after token mixing. GDConv applies a depthwise convolution over the entire feature map, enabling each channel to capture spatial importance with a small number of parameters while maintaining low computational overhead.

More specifically, in the embedding head, following the final encoding stage of FaceLiVTv2, the output feature map is first expanded to a 1284-channel representation using an efficient $1\times1$ point-wise convolution. Subsequently, a GDConv layer with a kernel size matching the spatial resolution of the feature map ($4\times4$) aggregates spatial information and reduces the feature map to a single vector representation. Finally, a linear projection aligns the resulting feature vector with the target 512-dimensional embedding space, as summarized in Table~\ref{tab:arch}. This design enables the GDConv head to learn adaptive, channel-wise spatial weighting while preserving the compactness and computational efficiency of the model, leading to more discriminative face embeddings compared to GAP-based aggregation.

\begin{figure}
    \centering
    \includegraphics[width=\columnwidth]{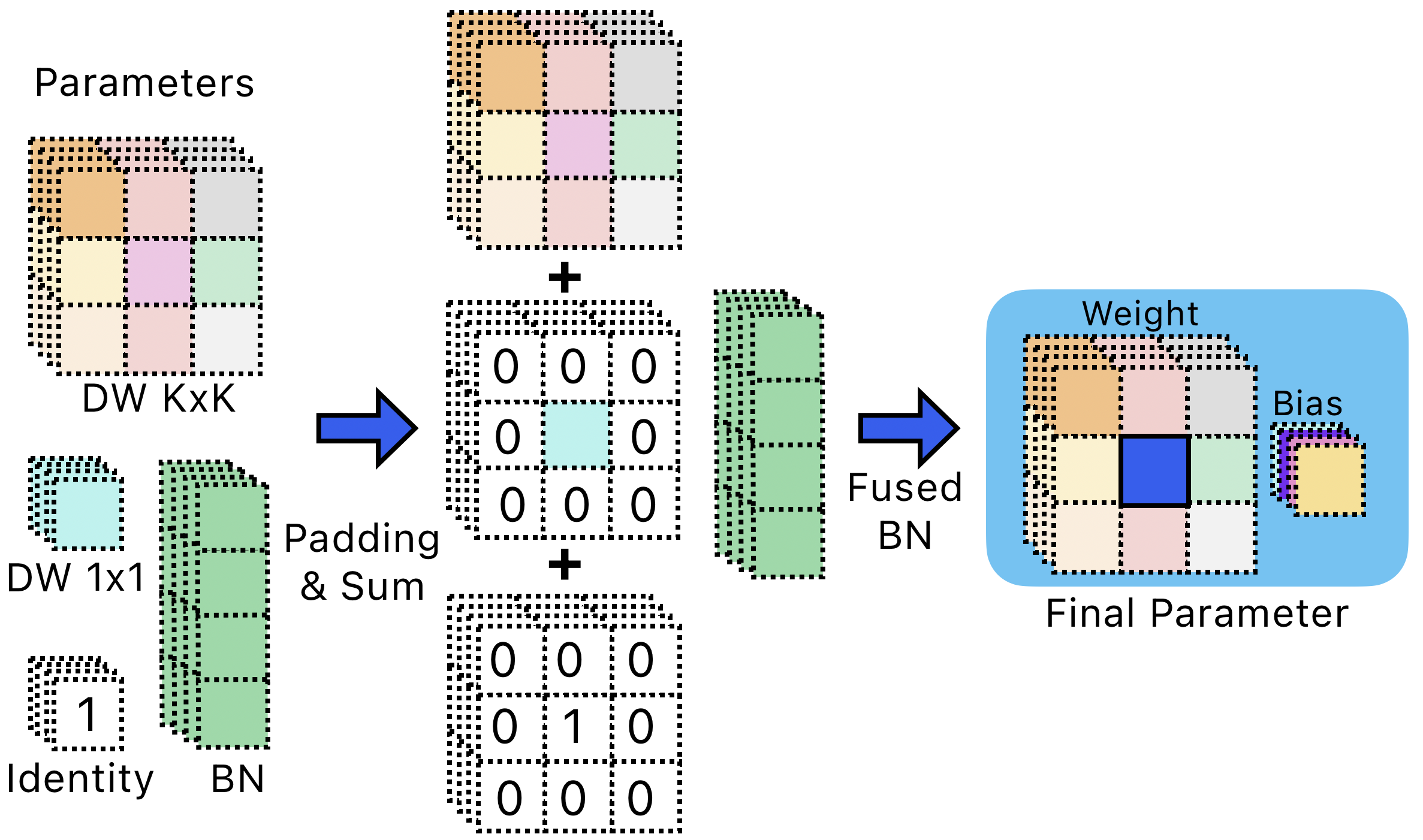}
    \caption{Structural Reparameterization on RepMix Block. This visualization illustrate the process of merging $DW_{3\times3}$, $DW_{1\times1}$, residual connection, and BN into single $DW_{3\times3}$. The residual connection is substituted by the identity weight with the dimension similar like $DW_{1\times1}$ weight.} 
    \label{fig:repmix}
\end{figure} 

\subsection{Reparameterized Token Mixer (RepMix)}
Following FaceLiVTv1 \cite{Setyawan2025FaceLiVT}, FaceLiVTv2 adopts the RepMix block for local feature extraction. RepMix is a lightweight convolutional token mixer designed to capture local spatial dependencies while remaining efficient. It combines a standard depthwise $3\times3$ convolution with an additional $1\times1$ depthwise convolution, followed by Batch Normalization (BN). During training, both branches contribute to feature learning. However, in the post-training process, they can be structurally reparameterized into a single equivalent convolution, thereby reducing latency without sacrificing accuracy.
Formally, given input $X_i$, RepMix computes the output feature in the training step as described in Eq. (\ref{eq:repmix}).
\begin{equation}
    X_i = X_i + \left\{BN(DW_{3\times 3}(X_i) + DW_{1\times 1}(X_i))\right\}.
    \label{eq:repmix}
\end{equation}
As illustrated in Fig. \ref{fig:repmix}, to reparameterize Eq. (\ref{eq:repmix}) in the deployment step, the $1\times 1$ depth-wise convolution and identity kernel weight, as a substitution for the residual connection, will be transformed into $3\times 3$ weight size through zero padding. Then it can be summed into a single $3\times 3$ weight convolution followed by BN, as described in Eq. (\ref{eq:bnconv}) , with $\mu$, $\sigma$, $\gamma$, and $\beta$ denoting mean, standard deviation, feature scale, and bias.
\vspace{-0.2\baselineskip}
\begin{equation}
    BN(DW_{3\times 3}(X_i))=\gamma\frac{(W*X_i+b)-\mu}{\sigma}+\beta.
    \label{eq:bnconv}
\end{equation}
In the final step, Eq. (\ref{eq:bnconv}) can be merged into a convolution layer only, as represented by Eq. (\ref{eq:fusedbn}): 
\vspace{-0.2\baselineskip}
\begin{equation}
    RepMix(X_i)=W' * X_i + b',
    \label{eq:fusedbn}
\end{equation}
where the transformed weight is $W' = W \frac{\gamma}{\sigma}$ and the adjusted bias is $b' = (b - \mu) \frac{\gamma}{\sigma} + \beta$. This simple yet effective design enriches local representation while enabling fast deployment on mobile devices. Hence, in the FaceLiVTv2, we extend RepMix in all stages and unify it with the Lite MHLA in the last two stages. 

\subsection{Lite Multi Head Linear Attention (Lite MHLA)}
In FaceLiVTv1 \cite{Setyawan2025FaceLiVT, setyawan2025facelivtapccas}, MHLA was proposed as a lightweight alternative to MHSA by replacing query--key interaction with linear projections. Given an input $X_i \in \mathbb{R}^{B \times C \times H \times W}$, 
it is reshaped into $X_i \in \mathbb{R}^{B \times C \times N}$ with $N = H \times W$, divided into $h$ heads, and computed as:
\begin{equation}
    \text{MHLA}(X_i) = \text{Concat}(LA_0, \ldots, LA_{He}),
    \label{eq:mhla}
\end{equation}
where each head is defined as in Eq.~\ref{eq:mhla_head}:
\begin{equation}
    LA_{He}(X_{He}) = W_o \left( \sigma(X_{He} \cdot W_i) \right),
    \label{eq:mhla_head}
\end{equation}
with $W_i \in \mathbb{R}^{N \times Nr}$ and $W_o \in \mathbb{R}^{Nr \times N}$ as learnable weight projections, where $Nr= N \times r$ denotes the projected dimension scaled by the $r$ ratio \cite{Setyawan2025FaceLiVT, setyawan2025facelivtapccas}. Despite its efficiency, the original MHLA introduced redundant linear projections across heads, while the two-layer MLP-style design (Linear-GELU-Linear) per head doubled the token interaction cost. Moreover, as demonstrated by \cite{touvron2022resmlp}, token-mixing can remain purely a linear layer without sacrificing representational power. Building on this principle, we introduce Lite MHLA, a simple activation-free linear mapping formulation that performs cross-token communication independently across multiple heads.

As shown in Fig. \ref{fig:v1_vs_v2}, Lite MHLA replaces the LayerNorm in MHLA with a single affine rescale transformation applied before head splitting. Since the input has already been normalized by the BN layer in the preceding RepMix block, a full re-normalization with mean and variance recomputation is redundant. A lightweight affine feature rescaling suffices to recalibrate the feature distribution and is particularly beneficial in mobile runtime, where normalization layers are disproportionately memory-bound (see Fig.~\ref{fig:breakdown}). 
This design demonstrates that a learnable affine feature rescaling operator stabilizes training in linear token-mixing architectures without sacrificing representational capacity (ablation in Sec.~IV-E confirms this empirically). Formally, the affine feature rescale transformation is defined as:
\begin{equation}
    \text{Aff}(X_i) = \alpha \odot X_i + \beta,
    \label{eq:affine}
\end{equation}
with $\alpha, \beta \in \mathbb{R}^{C \times 1}$ as learnable scale and bias parameters. It was applied both before and after the linear layer token interaction as channel-wise feature rescaling.

Each of the four heads, by default (see Sec.\ref{sec:abl_num_hed}), performs a single linear token interaction:
\begin{equation}
    \text{LiteMHLA}(X_i) = \text{Concat}(L^{lite}_0, \ldots, L^{lite}_{He}),
    \label{eq:litemhla}
\end{equation}
\begin{equation}
    L^{lite}_{He}(X_{He}) = \text{Aff}(X_{He}) \cdot W_i + B_i,
    \label{eq:litemhla_head}
\end{equation}
where $W_i \in \mathbb{R}^{N \times N}$, $B_i \in \mathbb{R}^{N \times 1}$, and $\text{Aff}(\cdot)$ are the affine transformations. Fig.~5 shows the PyTorch implementation of Eq.~(\ref{eq:litemhla}) and Eq.~(\ref{eq:litemhla_head}).

\textit{Remark on Complexity.} As summarized in Table~\ref{tab:complexity}, although Eq.~\ref{eq:litemhla_head} incurs $\mathcal{O}(N^2)$ cost per head per channel group, this remains practical because Lite MHLA eliminates the $Q/K/V$ and output projections ($4NC^2$), the attention product ($N^2C$), and the softmax normalization that MHSA requires, reducing the total complexity to $N^2C + \epsilon$ per block. Moreover, Lite MHLA operates exclusively at the small spatial resolutions of $7{\times}7$ ($N{=}49$) at stages~3 and $4{\times}4$ ($N{=}16$) at stages~4, where $N^2$ is inherently small and dominated by $C$ dimension. At stage~3 with $C{=}192$ (FaceLiVTv2-S), for example, the token interaction amounts to only ${\approx}0.46$\,M MAdds per block, which is negligible compared to the FFN (${4NC^2\approx}7.23$\,M).
\begin{table}[t!]
\centering \footnotesize
\setlength{\tabcolsep}{2pt} 
\caption{Complexity comparison of MHSA, MHLA, and Lite MHLA}
\label{tab:complexity}
\begin{tabular}{l|c|c|c}
\hline
\textbf{Component} & \textbf{MHSA} & \textbf{MHLA (v1)} & \textbf{Lite MHLA (v2)} \\
\hline
Q,K,V + Out Proj. & $4NC^2$ & -- & -- \\
Token Interaction & $2N^2C$  & $2(NrN)C$ & $N^2C$  \\
                  & ($QK^T$ and $Attn\cdot V$) & (MLP) & (Linear Proj.) \\
Nonlinearity & softmax & GELU & - \\
Normalization & LayerNorm & LayerNorm & Affine Transf. \\
\hline
\textbf{Total} & $4NC^2 + 2N^2C$ & $2(NrN)C$ & $\approx N^2C + \epsilon$ \\
\hline
\multicolumn{4}{l}{*$\epsilon$: affine rescale transformation overhead}
\end{tabular}%
\end{table}

\begin{figure}[t!]
\begin{center}
\begin{lstlisting}[language=Python]
class LiteMHLA(torch.nn.Module):
    def __init__(self, dim, resolution):
        super().__init__()
        self.n_head = 4
        self.res=resolution**2
        self.dim=dim
        linear=[]
        self.norm = Affine(dim) 
        for i in range(self.n_head):
            lin = nn.Linear(self.res, self.res)
            linear.append(lin)
        self.lin = torch.nn.ModuleList(linear)
        ls_init = torch.ones(dim).unsqueeze(-1)
        ls_init = ls_init.unsqueeze(-1)
        self.ls = nn.Parameter(1e-5 * ls_init)

    def forward(self, x):
        B,C,H,W = x.shape
        x = x.reshape(-1, self.dim, self.res)
        x = self.norm(x)
        x = list(x.chunk(self.n_head, dim=1))
        for i in range(self.n_head):
            x[i] = self.lin[i](x[i])
        x = torch.cat(x, dim=1)
        x =  self.ls * x.reshape(B,C,H,W)
        return x
\end{lstlisting}
\caption{Pytorch style code of Lite MHLA.}
\label{fig:mhlacode}
\end{center}
\end{figure}
In practice, Lite MHLA balances memory-bound and compute-bound operations by eliminating redundant Linear-Activation layer pairs and zero-feature normalization from merged BN and affine transformations. As shown in Fig.~\ref{fig:breakdown}, FaceLiVTv2 achieves nearly similar memory operation efficiency to GhostFaceNetV2-1~\cite{alansari2023ghostfacenets}, with significantly lower memory overhead than TransFace-S~\cite{dan2023transface} and KANFace-0.5~\cite{pham2025kanface}.

\section{Experiments}
\begin{table}[t!]
\setlength{\tabcolsep}{2.5pt} 
\caption{Architecture Details of FaceLiVTv2 Variants.}
\begin{center}
\begin{tabular}{llccccccc}
\hline
\multirow{2}{*}{Stage} & \multirow{2}{*}{Layer} &  \multirow{2}{*}{Res.}  & Kernel & \multirow{2}{*}{$n$} & \multicolumn{4}{c}{FaceLiVTv2 Channels}\\ \cline{6-9}
                       &           &   &  Size  &    & \textbf{XS} & \textbf{S} & \textbf{M} & \textbf{L} \\
\hline
Input & Image & $112^2$ & - & 1 & - & - & - & -\\
\hline
\multirow{2}{*}{Stem} & Conv & $56^2$ & 3$\times$3 & 1 & 16 & 24 & 28 & 32 \\
                      & Conv & $28^2$ & 3$\times$3 & 1 & 32 & 48 & 56 & 64 \\
\hline
\multirow{2}{*}{Stage-1}& RepMix Enc.  & $28^2$ & 3$\times$3 & 3 & 32 & 48 & 56 & 64 \\
                        & Downsampling & $14^2$ & 3$\times$3 & 1 & 64 & 96 & 112 & 128 \\
\hline
\multirow{2}{*}{Stage-2}& RepMix Enc.   & $14^2$ & 3$\times$3 & 3 & 64  & 96 & 112 & 128  \\
                        & Downsampling  & $7^2$  & 3$\times$3 & 1 & 128 & 192 & 224 & 256 \\
\hline
\multirow{2}{*}{Stage-3}& LiteMHLA Enc. & $7^2$ & 3$\times$3 & 9 & 128 & 192 & 224 & 256 \\
                        & Downsampling  & $4^2$ & 3$\times$3 & 1 & 256 & 384 & 448 & 512 \\
\hline
\multirow{2}{*}{Stage-4}& LiteMHLA Enc. & $4^2$ & 3$\times$3 & 3 & 256 & 384  & 448 & 512 \\ 
                        &  Conv         & $4^2$ & 1$\times$1 & 1 & \multicolumn{4}{c}{1284}  \\
\hline
\multirow{2}{*}{Head}& GDConv & $1^2$  & 4$\times$4 & 1 & \multicolumn{4}{c}{1284}\\
                     & Linear & $1^2$  &  - & 1&  \multicolumn{4}{c}{512}\\
\hline
\multicolumn{5}{c}{\textbf{Model MAdds}}          & 90 & 179  & 258  & 309 \\
\multicolumn{5}{c}{\textbf{Model Parameters (M)}}  & 2.9 & 4.62 & 7.04 & 8.52 \\
\hline 
\end{tabular}
\label{tab:arch}
\end{center}
\end{table}

\begin{table*}[!ht]
\centering
\setlength{\tabcolsep}{2.0pt}
\caption{Comparison of FaceLiVTv2 variant with SOTA on FR benchmarks. Mobile latency are measured on iPhone 15 Pro.}
\begin{tabular}{lcccccccccccccc}
\hline
\multirow{2}{*}{Model} & \multirow{2}{*}{Year} & Param & FLOPs & Training & \multirow{2}{*}{LFW$\uparrow$} & CA- & CP- & CFP & Age & Mean & \multicolumn{2}{c}{IJB (FAR@$10^{-4}$)$\uparrow$} & Latency \\ \cline{12-13}
& & (M)$\downarrow$ & (M)$\downarrow$ & Dataset & & LFW$\uparrow$ & LFW$\uparrow$ & -FP$\uparrow$ & DB30$\uparrow$ & Acc(\%)$\uparrow$ & B & C & ($ms$)$\downarrow$ \\ \hline
\multicolumn{15}{l}{\textit{Large-Scale FR Models (FLOPs $\geq$ 1G)}} \\
\hline
ResNet200-TopoFR~\cite{dan2024topofr} & '24 & 118.8 & 23.5G & Glint360K & 99.87 & - & - & 99.45 & 98.82 & - & 97.84 & 96.56 & 12.24 \\
TransFace-L~\cite{dan2025transface++, dan2023transface} & '25 & 271.6 & 25.4G & Glint360K & 99.85 & - & - & 99.37 & 98.66 & - & 96.73 & 97.85 & OOM \\
TransFace-B~\cite{dan2025transface++, dan2023transface} & '25 & 124.5 & 11.5G & Glint360K & 99.85 & - & - & 99.24 & 98.62 & - & 96.46 & 97.59 & 18.20 \\
TransFace-S~\cite{dan2025transface++, dan2023transface} & '25 & 86.7 & 5.8G & Glint360K & 99.85 & - & - & 98.96 & 98.56 & - & 96.12 & 97.45 & 14.31 \\
ResNet50-ArcFace~\cite{deng2019arcface} & '22 & 43.6 & 6.3G & Glint360K & 99.78 & - & - & 98.77 & 98.28 & - & 95.30 & 96.81 & 3.76 \\
SwiftFaceFormer-L3~\cite{luevano2024swiftfaceformer} & '24 & 28.0 & 2.0G & MS1MV3 & 99.75 & 96.03 & 90.70 & 97.80 & 97.55 & 96.36 & 92.92 & 94.70 & 2.59 \\
VarGFaceNet~\cite{yan2019vargfacenet} & '19 & 5.0 & 1.0G & MS1MV3 & 99.85 & 95.15 & 88.55 & 98.50 & 98.15 & 96.04 & 92.90 & 94.70 & 0.83 \\
PocketNetM256~\cite{boutros2022pocketnet} & '22 & 1.75 & 1.1G & CASIA-WF & 99.58 & 95.63 & 90.03 & 95.66 & 97.17 & 95.61 & 90.74 & 92.70 & 0.98 \\
PocketNetM128~\cite{boutros2022pocketnet} & '22 & 1.68 & 1.1G & CASIA-WF & 99.65 & 95.67 & 90.00 & 95.07 & 96.78 & 95.43 & 90.63 & 92.63 & 0.98 \\
\hline
\multicolumn{15}{l}{\textit{Lightweight FR Models (FLOPs $<$ 1G)}} \\
\hline
\rowcolor{gray!10}
KANFace-0.5~\cite{pham2025kanface} & '25 & 6.80 & 397 & WebFace12M & 99.82 & 95.48 & 92.65 & 98.31 & 96.90 & 96.63 & 93.69 & 95.64 & 9.98 \\
KANFace-0.5$\dagger$~\cite{pham2025kanface} & '25 & 6.80 & 397 & Glint360K & 99.75 & 95.52 & 91.90 & 97.35 & 97.21 & 96.35 & 93.26 & 94.34 & 9.98 \\
FaceLiVTv1-M~\cite{Setyawan2025FaceLiVT} & '25 & 9.8 & 386 & Glint360K & 99.70 & 95.76 & 90.97 & 97.20 & 97.60 & 96.25 & 93.70 & 95.70 & 0.67 \\
EdgeFace-S~\cite{george2024edgeface} & '24 & 3.6 & 306 & WebFace12M & 99.78 & 95.71 & 92.56 & 95.81 & 96.93 & 96.16 & 93.59 & 95.63 & 9.89 \\
EdgeFace-S$\dagger$~\cite{george2024edgeface} & '24 & 3.6 & 306 & Glint360K & 99.75 & 95.95 & 91.63 & 96.13 & 97.48 & 96.19 & 93.03 & 94.66 & 9.89 \\
SwiftFaceFormer-L1~\cite{luevano2024swiftfaceformer} & '24 & 11.8 & 805 & MS1MV3 & 99.68 & 95.80 & 90.10 & 96.61 & 96.95 & 95.83 & 91.81 & 93.82 & 1.20 \\
SwiftFaceFormer-S~\cite{luevano2024swiftfaceformer} & '24 & 6.0 & 485 & MS1MV3 & 99.60 & 95.78 & 90.00 & 96.49 & 96.83 & 95.74 & 91.56 & 93.54 & 0.65 \\
MobileFaceNet~\cite{chen2018mobilefacenets, martinez2021benchmarking} & '21 & 0.99 & 440 & MS1MV2 & 99.70 & 95.20 & 89.22 & 96.90 & 97.60 & 95.72 & 92.83 & 94.70 & 0.77 \\
ShuffleFaceNet-1.5~\cite{martindez2019shufflefacenet} & '21 & 2.6 & 577 & MS1MV2 & 99.67 & 95.05 & 88.50 & 97.26 & 97.32 & 95.56 & 92.30 & 94.30 & 0.69 \\
PocketNetS128~\cite{boutros2022pocketnet} & '22 & 0.92 & 587 & CASIA-WF & 99.58 & 95.48 & 88.63 & 94.21 & 96.10 & 94.80 & 89.44 & 91.62 & 0.88 \\
PocketNetS256~\cite{boutros2022pocketnet} & '22 & 0.99 & 587 & CASIA-WF & 99.66 & 95.50 & 88.93 & 93.34 & 96.36 & 94.76 & 89.31 & 91.33 & 0.88 \\
MixFaceNets-M~\cite{boutros2021mixfacenets} & '21 & 3.9 & 626 & MS1MV2 & 99.68 & - & - & - & 97.05 & - & 91.55 & 93.42 & 0.70 \\
\hline
\rowcolor{gray!15}
GhostFaceNetV1-1~\cite{alansari2023ghostfacenets} & '23 & 4.1 & 216 & MS1MV3 & 99.73 & 95.93 & 91.93 & 96.83 & 98.00 & 96.48 & 93.12 & 94.94 & 0.78 \\
KANFace-0.6~\cite{pham2025kanface} & '25 & 4.74 & 240 & WebFace12M & 99.65 & 95.32 & 91.47 & 97.17 & 95.52 & 95.83 & 92.95 & 94.75 & 6.52 \\
FaceLiVTv1-S~\cite{Setyawan2025FaceLiVT} & '25 & 5.89 & 237 & Glint360K & 99.70 & 95.63 & 90.70 & 95.10 & 96.60 & 95.55 & 91.20 & 92.70 & 0.47 \\
EdgeFace-XS~\cite{george2024edgeface} & '24 & 1.77 & 154 & WebFace12M & 99.73 & 95.28 & 91.82 & 94.37 & 96.00 & 95.44 & 92.67 & 94.85 & 5.82 \\
SwiftFaceFormer-XS~\cite{luevano2024swiftfaceformer} & '24 & 3.4 & 294 & MS1MV3 & 99.60 & 95.47 & 88.65 & 95.35 & 96.35  & 95.08 & 90.20 & 92.32 & 0.59 \\
\hline
\rowcolor{gray!10}
GhostFaceNetV2-2~\cite{alansari2023ghostfacenets} & '23 & 6.8 & 77 & MS1MV3 & 99.68 & 95.73 & 90.17 & 94.29 & 96.83 & 95.34 & 91.89 & 93.16 & 0.67 \\
GhostFaceNetV1-2~\cite{alansari2023ghostfacenets} & '23 & 4.1 & 60 & MS1MV3 & 99.68 & 95.60 & 90.07 & 93.31 & 96.92 & 95.12 & 91.25 & 93.45 & 0.60 \\
ShuffleFaceNet-0.5~\cite{martindez2019shufflefacenet} & '21 & 1.4 & 67 & MS1MV2 & 99.20 & - & - & 92.60 & 93.20 & - & - & - & 0.45 \\
\hline
\rowcolor{gray!10}
FaceLiVTv2-L & - & 8.52 & 309 & Glint360K & 99.80 & 96.00 & 93.07 & 98.26 & 98.02 & 97.03\textbf{(+0.40)} & 95.18\textbf{(+1.49)} & 96.59\textbf{(+0.95)} & 0.71\textbf{(14.0$\times\downarrow$)} \\
\hdashline
\rowcolor{gray!15}
FaceLiVTv2-M & - & 7.02 & 258 & Glint360K & 99.78 & 96.12 & 92.92 & 97.93 & 98.10 & 96.97\textbf{(+0.49)} & 95.02\textbf{(+1.90)} & 96.42\textbf{(+1.48)} & 0.65\textbf{(16.7\%$\downarrow$)} \\
\rowcolor{gray!15}
FaceLiVTv2-S & - & 4.62 & 179 & Glint360K & 99.78 & 95.93 & 92.45 & 97.47 & 97.82 & 96.69\textbf{(+0.21)} & 94.51\textbf{(+1.39)} & 95.99\textbf{(+1.05)} & 0.54\textbf{(30.8\%$\downarrow$)} \\
\hdashline
\rowcolor{gray!10}
FaceLiVTv2-XS & - & 2.9 & 90 & Glint360K & 99.63 & 95.58 & 90.38 & 95.23 & 96.68 & 95.50\textbf{(+0.16)} & 90.67\textbf{(-1.22)} & 91.25\textbf{(-1.91)} & 0.43\textbf{(35.8\%$\downarrow$)} \\
\hline
\multicolumn{15}{l}{$\dagger$ retrained from scratch using their official released PyTorch code under our experimental setup}  
\end{tabular}
\label{tab:benchmark}
\end{table*}

\subsection{Experiments Setup}
\subsubsection{Training}
We employed Glint360K\cite{an2021partial}, a large-scale public FR dataset that provides diverse and high-quality samples for robust model generalization. Glint360K is a recent and significantly larger dataset, comprising around 17 million face images from 360,232 unique identities. The dataset includes images from various demographics and capture conditions, making it suitable for training high-capacity models that generalize across unconstrained environments. For the low resolution face benchmark, we finetune the model with the TinyFace~\cite{cheng2018low} dataset. TinyFace \cite{cheng2018low} comprises 169,403 native low-resolution images of 5,139 identities, with a training subset containing 7,804 images of 2,570 identities.

The proposed FaceLiVTv2 was trained on both pre-aligned $112\times112$ resolution facial image datasets. They were transformed into tensors and normalized between -1 and 1. Training was performed over 50 epochs with a total batch size of 1026 on three Nvidia RTX A6000 GPUs in a distributed setting. The AdamW optimizer with a polynomial decay learning rate scheduler was used, with an initial learning rate of $6e^{-3}$ during the training process. We used l2 regularization with $l2=1e^{-2}$ to prevent overfitting. We also evaluated the model using two different loss functions: CosFace \cite{wang2018cosface} and ArcFace \cite{deng2019arcface}, with a 512-dimensional embedding size. For TinyFace \cite{cheng2018low}, we fine-tuned the model for 40 epochs with a batch size of 8 using the AdamW optimizer with an initial learning rate of $2e^{-4}$, without any distillation or restoration.

\subsubsection{Test} 
The performance of the FaceLiVTv2 model proposed in this study was assessed using eight distinct benchmark datasets. These datasets, chosen for evaluation purposes, include Labeled Faces in the Wild (LFW) \cite{huang2008labeled}, Cross-age LFW (CA-LFW) \cite{zheng2017cross}, Cross Pose LFW (CP-LFW) \cite{zheng2018cross}, Celebrities in Frontal-Profile in the Wild (CFP-FP) \cite{sengupta2016frontal}, AgeDB-30 \cite{moschoglou2017agedb}, IARPA Janus Benchmark-B (IJB-B) \cite{whitelam2017iarpa}, IARPA Janus Benchmark-C (IJB-C) \cite{maze2018iarpa}, and TinyFace \cite{cheng2018low} for low-resolution face test protocol.
\begin{figure}[t!]
    \centering
    \includegraphics[width=\columnwidth]{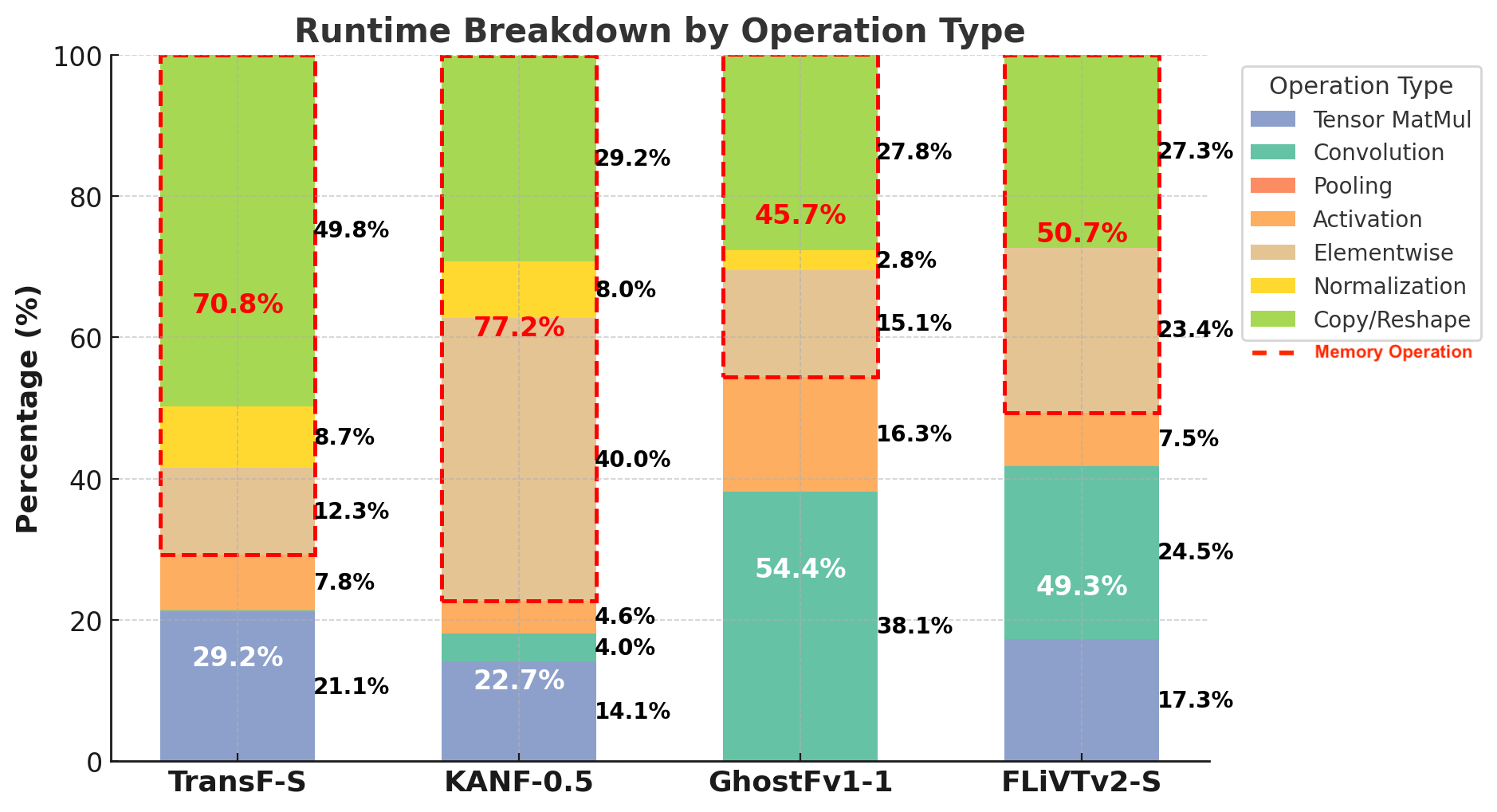}
    \caption{Model in CoreML runtime breakdown. Operations highlighted in the redbox represent memory-bound operations, where memory access dominates execution time over computation.}
    \label{fig:breakdown}
\end{figure}
We further assessed model efficiency not only in terms of FLOPs but also through an evaluation of inference speed. The inference speed test was carried out on various computing devices, such as iPhone 15 Pro for mobile, Jetson AGX Orin for the edge platform, Intel i5-12500 with 64 GB of RAM for consumer-grade CPU, and RTX-5090 for GPU. For the mobile latency evaluation, all models were converted to Core ML using coremltools based on the official implementations, and latency was measured on the iPhone 15 Pro running iOS 26. For the edge device, CPU, and GPU, all models were benchmarked using ONNX Runtime as an open standard for machine learning interoperability. The detailed conversion process from PyTorch to Core ML and ONNX is provided in code and the supplementary material.

\subsection{Comparison With SOTA}
Table \ref{tab:benchmark} shows the comparison of our new FaceLiVTv2 with SOTA lightweight face
recognition models in the literature across different benchmarking datasets. We categorize the model based on FLOPs into under 100 MFLOPs, $100\leq$ MFLOPs $<300$, $300\leq$ MFLOPs $<1024$, and $\geq1024$; then we sort it by mean accuracy among LFW, CA-LFW, CP-LFW, CFP-FP, and AgeDB-30.
Starting from the most compact configuration under 100 MFLOPs, FaceLiVTv2-XS demonstrates that meaningful FR performance can be achieved under extremely tight computational constraints. With only 2.9 M parameters and 90 MFLOPs, FaceLiVTv2-XS attains 0.16\% higher average accuracy across 5 benchmarks while delivering a 35.8\% reduction in mobile latency compared with GhostFaceNetV2-2\cite{alansari2023ghostfacenets}. Although its accuracy on IJB-B and IJB-C is 1.22\% to 1.91\% lower, the substantial latency savings make FaceLiVTv2-XS particularly suitable for ultra-edge and always-on mobile scenarios where responsiveness and energy efficiency are prioritized over absolute peak accuracy.

Moving to MFLOPs ranging from 100 to 300, FaceLiVTv2-S achieves a favorable balance between recognition accuracy and computational efficiency relative to recent lightweight and hybrid face-recognition networks. Although the proposed model contains only 4.62 M parameters and 179 MFLOPs, it consistently delivers superior results across diverse benchmarks, with strong cross-age robustness of 97.82\% on AgeDB-30, cross-pose robustness of 92.45\% on CP-LFW, and high template-based verification accuracy of 94.51\% on IJB-B and 95.99\% on IJB-C. Compared with CNN-based models, such as GhostFaceNetV2-2~\cite{alansari2023ghostfacenets}, FaceLiVTv2-S achieves higher accuracy with mean accuracy increases of 0.21 while maintaining 30.8\% faster mobile inference. It demonstrates strong generalization capability, which is critical for real-world deployments.

As model capacity increases, FaceLiVTv2-M continues this trend by outperforming lightweight FR models in the 100–300 MFLOPs range across all benchmarks, achieving a mean accuracy of 96.97\% on 5 FR benchmarks, 95.02\% on IJB-B, and 96.42\% on IJB-C. Despite its higher representational capacity, FaceLiVTv2-M maintains a favorable efficiency profile, remaining 16.7\% faster than GhostFaceNetV1-1 \cite{alansari2023ghostfacenets} and consistently surpassing the EdgeFace–KANFace line of lightweight hybrid models \cite{george2024edgeface, pham2025kanface} in terms of both recognition accuracy and inference latency.

At the highest model size, FaceLiVTv2-L reaches the strongest overall performance, achieving a mean accuracy of 97.03\% with 95.18\% on IJB-B and 96.59\% on IJB-C while preserving low mobile inference latency. Compared with SOTA models with FLOPs ranging from 300M to 1G, FaceLiVTv2-L consistently outperforms its predecessor FaceLiVTv1-M~\cite{Setyawan2025FaceLiVT} with a similar inference latency. Moreover, it achieves a mean accuracy that is 0.4 to 0.84 higher than KANFace-0.5~\cite{pham2025kanface} and EdgeFace-S~\cite{george2024edgeface} on most benchmarks, which were trained on the WebFace12M and Glint360K datasets with $\approx$14$\times$ faster mobile inference. 

Compared with large scale FR models that were trained on similar datasets such as ResNet200-TopoFR\cite{dan2024topofr} and TransFace variants~\cite{dan2025transface++}, FaceLiVTv2-L narrows the mean accuracy gap on LFW, CFP-FP, and AgeDB-30 to only 0.69\% while requiring $13.9\times$ fewer parameters and $76.1\times$ fewer FLOPs than ResNet200-TopoFR\cite{dan2024topofr}. Remarkably, FaceLiVTv2-L marginally surpasses ResNet200-TopoFR by 0.03\% on IJB-C, demonstrating competitive open-set verification capability. Against TransFace-L~\cite{dan2025transface++} as the strongest Transformer-based model, FaceLiVTv2-L achieves a mean accuracy gap of only $0.60\%$ on LFW, CFP-FP, and AgeDB-30, while using $31.9\times$ fewer parameters and $82.2\times$ fewer FLOPs. Compared with large scale mobile FR models, FaceLiVTv2-L remains $3.65\times$ faster than SwiftFaceFormer-L3~\cite{luevano2024swiftfaceformer}, 14\% faster than VarGFaceNet~\cite{yan2019vargfacenet} while achieving higher accuracy. This demonstrates that FaceLiVTv2-L offers a compelling alternative to heavier architectures, combining strong generalization with deployment-friendly efficiency.

\begin{table}[t!]
\centering
\setlength{\tabcolsep}{2.5pt} 
\caption{Cross-Platform Latency Comparison of FaceLiVTv2 with SOTA}
\begin{tabular}{lcccc} 
\hline
\multirow{2}{*}{Model} & Mean & \multicolumn{3}{c}{ONNX Latency (ms) $\downarrow$} \\ \cline{3-5} 
                       & Acc(\%)$\uparrow$ & GPU & CPU & Jetson-(GPU/CPU) \\ \hline   
\rowcolor{gray!10}
KANFace-0.5~\cite{pham2025kanface}       & 96.63 & 1.96 & 6.19 & 7.12 / 20.47  \\
FaceLiVTv1-M~\cite{Setyawan2025FaceLiVT,setyawan2025facelivtapccas} & 96.25 & 1.83 & 3.61 & 7.04 / 13.65 \\
EdgeFace-S~\cite{george2024edgeface}    & 96.16 & 1.84 & 2.90 & 6.85 / 11.16 \\
SwiftFaceFormer-L1~\cite{luevano2024swiftfaceformer} & 95.83 & 1.36 & 4.51 & 5.53 / 10.85 \\
SwiftFaceFormer-S~\cite{luevano2024swiftfaceformer} & 95.74 & 1.28 & 2.53 & 5.28 / 8.18 \\
\hline
\rowcolor{gray!15}
GhostFaceNetV1-1~\cite{alansari2023ghostfacenets} & 96.48 & 1.12 & 4.53 & 5.08 / 19.68 \\
KANFace-0.6~\cite{pham2025kanface}                & 95.83 & 1.92 & 4.01 & 6.60 / 14.37 \\
FaceLiVTv1-S~\cite{Setyawan2025FaceLiVT,setyawan2025facelivtapccas} & 95.55 & 1.73 & 2.10 & 6.40 / 8.02 \\
EdgeFace-XS~\cite{george2024edgeface}             & 95.44 & 1.83 & 2.12 & 6.42 / 8.04 \\
SwiftFaceFormer-XS~\cite{luevano2024swiftfaceformer} & 95.08 & 1.17 & 1.82 & 4.54 / 6.20\\
\hline
\rowcolor{gray!10}
GhostFaceNetV2-2~\cite{alansari2023ghostfacenets} & 95.34 & 1.49 & 2.80 & 6.89 / 12.02  \\
GhostFaceNetV1-2~\cite{alansari2023ghostfacenets} & 95.12 & 1.03 & 2.21 & 4.96 / 8.32  \\
\hline 
\rowcolor{gray!10}
FaceLiVTv2-L &97.03 & 1.29 & 2.40 & 5.45 / 8.80 \\ \rowcolor{gray!15}
            & \textbf{(+0.40)} & \textbf{(34\%)$\downarrow$} & \textbf{(47\%)$\downarrow$} & \textbf{(21\%)$\downarrow$} / \textbf{(57\%)$\downarrow$} \\ \hdashline
\rowcolor{gray!15} 
FaceLiVTv2-M & 96.97&1.27&2.08& 5.39 / 7.61 \\ \rowcolor{gray!15}
            & \textbf{(+0.49)} & \textbf{(12\%)$\uparrow$} & \textbf{(23\%)$\downarrow$} & \textbf{(2\%)$\downarrow$} / \textbf{(60\%)$\downarrow$} \\
\rowcolor{gray!15}
FaceLiVTv2-S & 96.69 & 1.25 & 1.64 & 5.35 / 6.42  \\ \rowcolor{gray!15} 
            & \textbf{(+0.21)} & \textbf{(10\%)$\uparrow$} & \textbf{(66\%)$\downarrow$} & \textbf{(4\%)$\downarrow$} / \textbf{(67\%)$\downarrow$} \\
\hdashline \rowcolor{gray!10} 
FaceLiVTv2-XS & 95.50 & 1.23 & 1.24 & 5.08 / 4.57 \\ \rowcolor{gray!15} 
            & \textbf{(+0.16)} & \textbf{(17\%)$\downarrow$} & \textbf{(43\%)$\downarrow$} & \textbf{(25\%)$\downarrow$} / \textbf{(62\%)$\downarrow$} \\
\hline
\end{tabular}
\label{tab:cross_benchmark}
\end{table}

\subsection{Cross-Platform Inference Latency Analysis}
Across all evaluated platforms in Table~\ref{tab:cross_benchmark}, FaceLiVTv2 demonstrates a strong accuracy–latency trade-off, indicating robust inference generalization. On Jetson AGX Orin, FaceLiVTv2-S and FaceLiVTv2-M achieve 2\% to 4\% lower latency than GhostFaceNetV1-1, while FaceLiVTv2-XS is 25\% faster than GhostFaceNetV2-2. FaceLiVTv2-L is 21\% faster than KANFace-0.5, with a 0.40 improvement in mean accuracy. On Jetson CPU, FaceLiVTv2 delivers 57–67\% faster inference than competing models within the same MFLOPs range, underscoring its suitability for real-time edge deployment.

\begin{figure}[t!]
    \centering
    \includegraphics[width=\columnwidth]{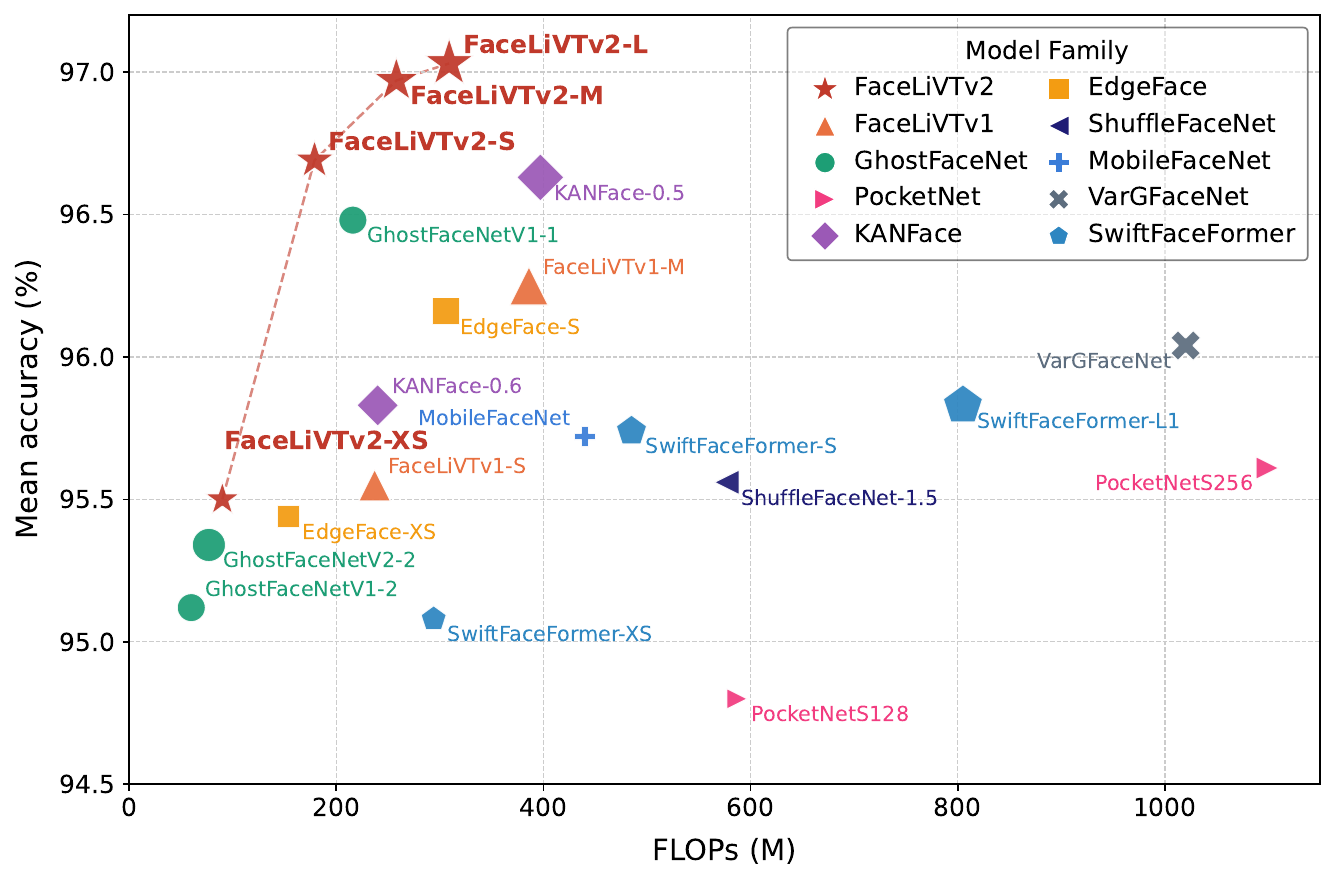}
    \includegraphics[width=\columnwidth]{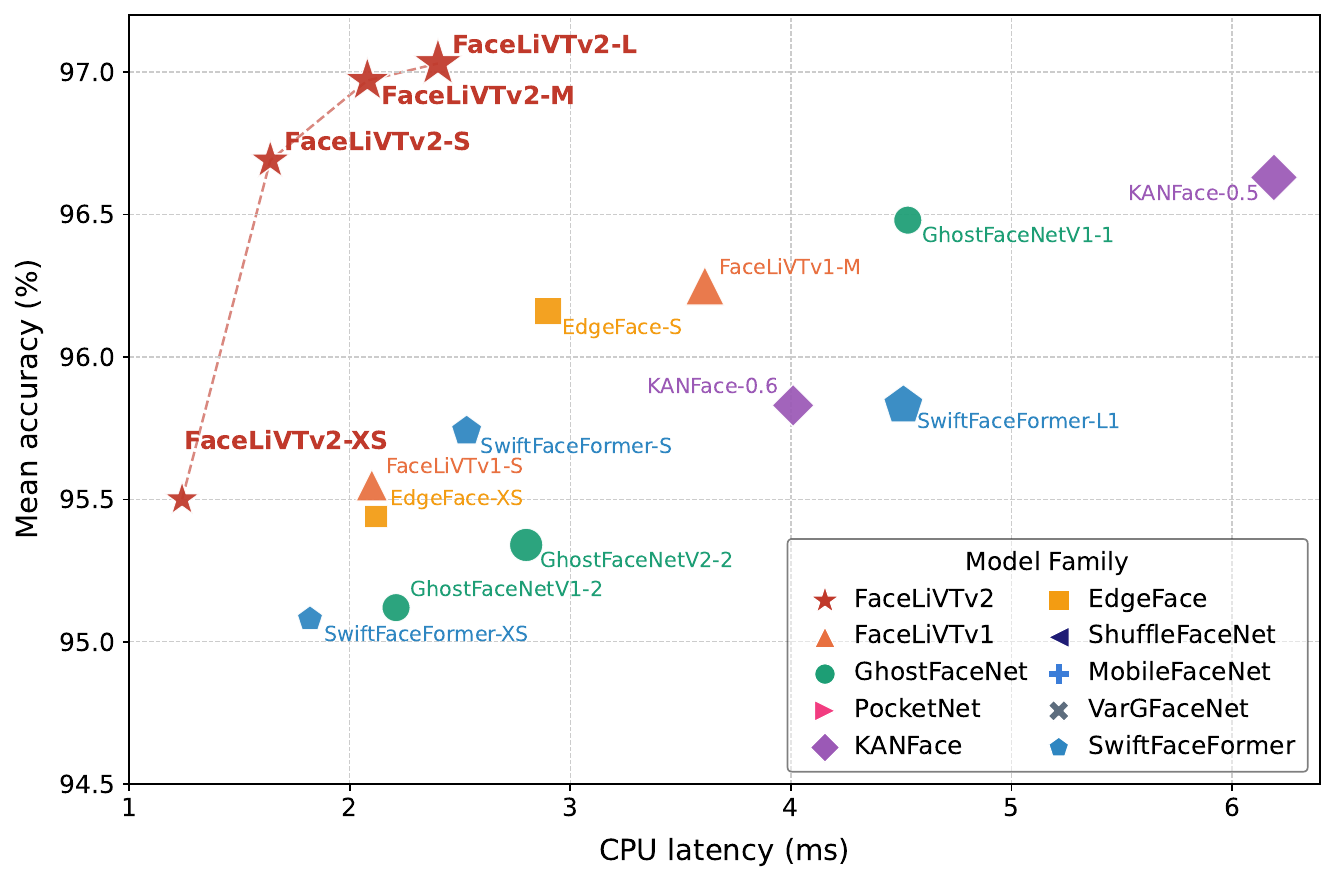}
    \caption{Comparison of our proposed FaceLiVTv2 with other lightweight FR models on mean accuracy of FR benchmark dataset v.s. FLOPs and mean accuracy v.s i5-12500 CPU Latency. }
    \label{fig:pareto_flops_cpu}
\end{figure}

On the CPU and GPU platforms, FaceLiVTv2 continues to exhibit favorable efficiency characteristics. On CPU, FaceLiVTv2-S and FaceLiVTv2-M reduce inference latency by 66\% to 23\%, respectively, compared with GhostFaceNetV1-1\cite{alansari2023ghostfacenets}, while FaceLiVTv2-L consistently outperforms KANFace\cite{pham2025kanface} and EdgeFace\cite{george2024edgeface} in both accuracy and runtime, as shown in Fig. \ref{fig:pareto_flops_cpu}. On GPU, FaceLiVTv2 variants maintain competitive lower inference latency times over other lightweight transformer models such as EdgeFace\cite{george2024edgeface}, KANFace\cite{pham2025kanface}, and SwiftFaceFormer\cite{luevano2024swiftfaceformer}, although they are 13–20\% slower than the pure CNN GhostFaceNetV1\cite{alansari2023ghostfacenets}, which is highly optimized for GPU execution.

\begin{table}[t!]
\centering
\setlength{\tabcolsep}{2.5pt} 
\renewcommand{\arraystretch}{1.05} 
\caption{Comparison of FaceLiVTv2 Variant with SOTA on TinyFace}
\begin{tabular}{lcccccc} 
\hline
\multirow{2}{*}{Model} &  FLOP & \multicolumn{4}{c}{Rank} \\ \cline{3-6}
      &   (M)$\downarrow$    &  1     &    5   &   20    &   50    \\ \hline
VGGFace \cite{khalid2020resolution}   &  -  & 30.4 & - & 40.4 & 42.7 \\
CenterFace \cite{khalid2020resolution} &  -  & 32.1 & - & 44.5 & 48.4 \\
MobileFaceNet \cite{martinez2021benchmarking} from \cite{zhang2023texture}     & 440 & 42.3 & 44.6 & 47.1 & 50.2 \\  
ShuffleFaceNet-1.5 \cite{martindez2019shufflefacenet, martinez2021benchmarking} from \cite{zhang2023texture} & 577 & 50.2 & 52.4 & 54.5 & 55.3 \\
VarGFaceNet \cite{yan2019vargfacenet} from \cite{zhang2023texture}  & 1022 & 52.8 & 54.1 & 56.3 & 58.2 \\
ResNet34 \cite{zhang2023texture}  & 8G & 56.1 & 56.9 & 59.0 & 63.8 \\
CSRI\cite{kolouri2015transport}   & -  & 44.8 & -    & 60.4 & 65.1 \\ 
ResNet50 \cite{zhang2023texture}  & 12G & 61.2 & 63.7 & 65.9 & 68.2 \\ 
RI-ResNet34 \cite{khalid2020resolution} & 8G & 70.4 & - & 82.2 & 85.4 \\ \hline
FaceLiVTv2-XS   & 90 & 46.1 & 52.7 & 58.0 & 61.7  \\
FaceLiVTv2-S   & 179 & 47.8 & 55.5 & 60.5 & 64.0  \\
FaceLiVTv2-M   & 258 & 49.6 & 56.3 & 60.8 & 64.2  \\
FaceLiVTv2-L   & 309 & 49.5 & 56.3 & 61.7 & 65.0  \\
\hline
\end{tabular}
\label{tab:tinybenchmark}
\end{table}

\subsection{Low-Resolution Face Benchmark Analysis}
Table \ref{tab:tinybenchmark} evaluates FaceLiVTv2 on the low-resolution TinyFace\cite{cheng2018low} benchmark, which is widely regarded as a challenging testbed for surveillance-style FR due to severe resolution degradation, blur, and noise. Despite its compact design, FaceLiVTv2-XS achieves 46.1\% Rank-1 and 61.7\% Rank-50 accuracy with only 90 MFLOPs, outperforming MobileFaceNet \cite{martinez2021benchmarking} and several classical baselines while operating at substantially lower computational costs. As model capacity increases, FaceLiVTv2 exhibits a clear and consistent performance gain: FaceLiVTv2-S improves Rank-1 accuracy to 47.8\%, while FaceLiVTv2-M and FaceLiVTv2-L further raise Rank-1 accuracy to 49.6\% and 49.5\%, respectively. Notably, FaceLiVTv2-L achieves a Rank-50 accuracy of 65.0\%, surpassing ShuffleFaceNet-1.5 \cite{martindez2019shufflefacenet, martinez2021benchmarking}, VarGFaceNet \cite{yan2019vargfacenet}, and ResNet34\cite{zhang2023texture}, while requiring orders of magnitude fewer FLOPs. These results demonstrate that FaceLiVTv2 maintains robust recognition performance under extremely low-resolution conditions with only finetuning, without any special techniques such as distillation or image restoration~\cite{zhang2023texture, khalid2020resolution}, indicating improved generalization to surveillance and cross-resolution scenarios despite its lightweight architecture.
\begin{table*}[t!]
\centering
\setlength{\tabcolsep}{2.5pt} 
\renewcommand{\arraystretch}{1.1} 
\caption{Ablation On The Overall Impact Of Components On FaceLiVTv2-S Variant with Glint360K as a Training Dataset}
\begin{tabular}{>{\centering}llllllllll } 
\hline
Architectural  & Param & FLOPs &LFW & CA-LFW &CP-LFW & CFP-FP & AgeDB-30   & Mean Acc   & Latency \\ 
 Ablation      & (M)$\downarrow$ & (M)$\downarrow$ & (\%)$\uparrow$ & (\%)$\uparrow$ & (\%)$\uparrow$ & (\%)$\uparrow$ & (\%)$\uparrow$ & (\%)$\uparrow$ & (ms)$\downarrow$ \\
\hline
FaceLiVTv1-S (Baseline) & 5.9 & 237 & 99.70 & 95.63 & 90.70 & 95.10 & 96.60 & 95.55 & 0.47 \\
Adjust Dim$\uparrow$, Depth$\uparrow$, \& MLP$\downarrow$  & 6.5\textbf{(10\%)$\uparrow$} & 203\textbf{(14\%)$\downarrow$} & 99.70\textbf{(+0.00)} & 95.63\textbf{(+0.00)} & 91.53\textbf{(+0.83)} & 95.73\textbf{(+0.63)} & 97.05\textbf{(+0.45)} & 95.92\textbf{(+0.37)} & 0.67\textbf{(42\%)$\uparrow$} \\
 \hline
GAP $\rightarrow$ GDConv Recog. Head & 7.4\textbf{(25\%)$\uparrow$} & 210\textbf{(11\%)$\downarrow$} & 99.72\textbf{(+0.02)} & 95.75\textbf{(+0.12)} & 91.20\textbf{(+0.50)} & 96.12\textbf{(+1.02)} & 97.03\textbf{(+0.43)} & 95.96\textbf{(+0.40)} & 0.73\textbf{(55\%)$\uparrow$} \\
Separate $\rightarrow$ Combined RepMix & 7.4\textbf{(25\%)$\uparrow$} & 211\textbf{(11\%)$\downarrow$} & 99.73\textbf{(+0.03)} & 95.87\textbf{(+0.24)} & 91.22\textbf{(+0.52)} & 96.15\textbf{(+1.05)} & 97.23\textbf{(+0.63)} & 96.00\textbf{(+0.45)} & 0.74\textbf{(57\%)$\uparrow$} \\ 
\rowcolor{gray!20}
MHLA $\rightarrow$ LiteMHLA  & 4.6\textbf{(22\%)$\downarrow$} & 179\textbf{(24\%)$\downarrow$} & 99.78\textbf{(+0.08)} & 95.93\textbf{(+0.30)} & 92.45\textbf{(+1.75)} & 97.47\textbf{(+2.37)} & 97.82\textbf{(+1.22)} & 96.69\textbf{(+1.14)} & 0.52\textbf{(10\%)$\uparrow$} 
\\ \hline
FaceLiVTv1-M & 9.8 & 386 & 99.70 & 95.76 & 90.97 & 97.20 & 97.60 & 96.25 & 0.67 \\
FaceLiVTv2-S & 4.6\textbf{(53\%)$\downarrow$} & 179\textbf{(53\%)$\downarrow$} & 99.78\textbf{(+0.08)} & 95.93\textbf{(+0.27)} & 92.45\textbf{(+1.48)} & 97.47\textbf{(+0.27)} & 97.82\textbf{(+0.22)} & 96.69\textbf{(+0.44)} & 0.52\textbf{(22\%)$\downarrow$} \\
\hline
\end{tabular}
\label{tab:overall_abl}
\end{table*}
\begin{table}[t!]
\centering 
\setlength{\tabcolsep}{2.5pt} 
\caption{Block Configurations Ablation of FaceLiVTv2-S}
\begin{tabular}{ccccccccc} 
 \hline
Block Config. & Param & FLOPs & \multirow{2}{*}{LFW} & CA &CP & CFP & Age & Lat \\ 
{[4-Stages]}    & (M) & (M) &  & LFW & LFW & FP &DB30  & (ms) \\ \hline
{[R, R, R, R]} & 4.52 & 175 & 99.73 & 95.88 & 92.07 & 97.19 & 97.67 & 0.50 \\
\hline
\rowcolor{gray!20}
{[R, R, RL, RL]} & 4.62 & 179 & 99.78 & 95.93 & 92.45 & 97.47 & 97.82 & 0.54 \\
\hline
{[RL, RL, RL, RL]} & 12.47 & 279 & 99.80 & 95.99 & 92.54 & 97.57 & 97.83 & 1.36 \\
\hline
\end{tabular}
\label{tab:block_config_abl}
\end{table}
\subsection{Ablation Study} 
We conducted several ablation studies, including the RepMix design, loss function, Lite MHLA design, and overall architectural changes. We used Glint360K with a similar training setup across all ablations.
\subsubsection{Ablation on Overall Impact of Architectural Change}
Table~\ref{tab:overall_abl} presents the cumulative impact of each architectural change on FaceLiVTv2-S, starting from the FaceLiVTv1-S baseline trained on Glint360K. Adjusting the stage dimensions from $[40, 80, 120, 320]$ to $[48,96,192,320]$ and depths from $[2,4,6,2]$ to $[3,3,9,3]$ with an MLP expansion ratio from 3 to 2 yields a 42\% latency increase (from 0.47\,ms to 0.67\,ms) but provides meaningful accuracy gains across all benchmarks, most notably +0.83\% on CP-LFW
and +0.45\% on AgeDB-30, establishing a stronger feature backbone for subsequent modifications. Replacing global average pooling with a global depthwise convolution recognition head further improves
accuracy--particularly on CFP-FP and AgeDB-30--while only marginally increasing latency to 0.73\,ms, confirming that adaptive spatial aggregation captures more discriminative information than uniform pooling. Transitioning from separate RepMix and global mixer stages to the combined RepMix block yields a mean accuracy gain of 0.45 from the baseline, with a similar latency cost profile, indicating that tighter local-global coordination benefits representation quality.

The most impactful change is the replacement of MHLA with Lite MHLA. This change reduces the number of parameters by 22\% and FLOPs by 24\% from the baseline while simultaneously achieving the largest accuracy improvements across all benchmarks: +1.75\% on CP-LFW, +2.37\% on CFP-FP, +1.22\% on AgeDB-30, and +1.14\% on the overall mean. The latency also drops sharply from 0.74\,ms to 0.52\,ms when compared with the previous version of MHLA. This confirms that the static linear token interaction in Lite MHLA is not only more parameter-efficient but also more effective than the two-layer MLP design in the original MHLA, as eliminating the intermediate nonlinearity and redundant projections allows the network to learn cleaner token-to-token relationships.

The final FaceLiVTv2-S configuration achieves 96.69\% mean accuracy with only 4.6\,M parameters and 0.52\,ms latency, outperforming the significantly larger FaceLiVTv1-M (53\% larger than FaceLiVTv2-S) on every benchmark while being 22\% faster. This demonstrates that the architectural refinements in FaceLiVTv2 are complementary and collectively yield a model that is both smaller and more accurate than its predecessor, validating the co-design of Lite MHLA, combined RepMix, and adaptive spatial aggregation as an effective strategy for mobile FR.

\textit{Block Configuration Analysis.} Table~\ref{tab:block_config_abl} examines how placing Lite MHLA across stages affects the accuracy-efficiency trade-off, where R and RL denote RepMix-only and combined RepMix-Lite MHLA stages, respectively. Using RepMix alone ([R, R, R, R]) yields the most compact model but lags on challenging benchmarks like CFP-FP and CP-LFW, where global context is essential for handling large pose and age variations. Adding Lite MHLA in the last two stages ([R, R, RL, RL]) costs only 0.10\,M extra parameters and 0.04\,ms latency, yet it improves CFP-FP by +0.28\% and CP-LFW by +0.38\%, confirming that global token interaction is most beneficial at later stages where spatial resolution is small and the computational cost of Lite MHLA is negligible. Conversely, applying Lite MHLA at every stage nearly triples the parameters to 12.47\,M and increases latency by 2.5$\times$ for only marginal accuracy gains, indicating that early high-resolution stages mainly capture low-level spatial patterns, where global token interaction offers limited benefits and introduces unnecessary overhead. The [R, R, RL, RL] configuration thus represents the optimal trade-off, concentrating global modeling where it matters most.

\begin{table}[t!]
\centering 
\setlength{\tabcolsep}{4.5pt} 
\renewcommand{\arraystretch}{1.05} 
\caption{Loss Function Ablation of FaceLiVTv2-S}
\begin{tabular}{lccccc } 
\hline
\multirow{2}{*}{Loss Function}  & \multirow{2}{*}{LFW} & CA &CP & CFP & Age   \\ 
                  &  & LFW & LFW & FP &DB30   \\ \hline
ArcFace\cite{deng2019arcface} & 99.75 & 95.75 & 91.63 & 96.63 & 97.65  \\
\rowcolor{gray!20}
CosFace\cite{wang2018cosface} & 99.78 & 95.93 & 92.45 & 97.47 & 97.82 \\ 
\hline
\end{tabular}
\label{tab:loss_abl}
\end{table}

\subsubsection{Ablation of Loss Function}
To examine the influence of different margin-based loss functions on the discriminative capacity of FaceLiVTv2, we evaluated two widely used angular-margin formulations; specifically, we trained the FaceLiVTv2-S variant with ArcFace \cite{deng2019arcface} and CosFace \cite{wang2018cosface}. As shown in Table \ref{tab:loss_abl}, both loss functions yield strong performance across all benchmarks, demonstrating the robustness of the proposed architecture to the choice of supervision strategy. When trained with ArcFace, the model achieves 99.75\% on LFW and 97.63\% on AgeDB-30, confirming stable intra-class compactness and inter-class separability. However, CosFace provides a consistent improvement across nearly all datasets, notably on CP-LFW and CFP-FP, where accuracy gains are from 91.63\% to 92.45\% and from 96.63\% to 97.47\%, respectively. 

The superior performance of CosFace can be attributed to its additive cosine-margin formulation, which enforces a constant angular margin in the normalized feature space, promoting clearer class boundaries and faster convergence. In contrast, ArcFace applies an additive angular margin, which slightly increases the decision boundary curvature and may require more careful margin scaling. Overall, CosFace proves to be more stable for training lightweight architectures such as FaceLiVTv2-S, providing improved generalization and better alignment between feature embeddings and classification weights.
\subsubsection{Ablation of RepMix}
To evaluate the effectiveness of RepMix, we conduct an ablation study by removing the residual and BN reparameterization components. In addition, we analyze the multiscale kernel contribution of the $DW_{1\times1}$ block in supporting the main $DW_{3\times3}$ operation. Table~\ref{tab:repmix_abl} summarizes the effects of these design changes. The results indicate that both residual and BN reparameterization significantly influence latency. Specifically, removing residual reparameterization increases latency from 0.54~ms to 0.63~ms, primarily due to the additional memory access introduced by the residual path. Similarly, not fusing BN leads to a latency increase from 0.54~ms to 0.62~ms, accompanied by a rise in FLOPs. Furthermore, the $DW_{1\times1}$ block plays a crucial role in stabilizing training by providing an additional gradient propagation path. Eliminating this block results in a noticeable reduction in accuracy across nearly all benchmark tests. 
\begin{table}[t!]
\centering 
\setlength{\tabcolsep}{2.5pt} 
\caption{RepMix Block Ablations of FaceLiVTv2-S}
\begin{tabular}{lcccccccc} 
\hline
\multirow{2}{*}{Ablations} & Param & FLOPs & \multirow{2}{*}{LFW} & CA &CP & CFP & Age & Lat \\ 
    & (M) & (M) &  & LFW & LFW & FP &DB30  & (ms) \\ \hline
\rowcolor{gray!20}
Baseline & 4.62 & 179 & 99.78 & 95.93 & 92.45 & 97.47 & 97.82 & 0.54 \\
\hline
w/o Res Rep  & 4.62 & 179 & 99.78 & 95.93 & 92.45 & 97.47 & 97.82 & 0.63 \\
w/o fused BN & 4.65 & 183 & 99.78 & 95.93 & 92.45 & 97.47 & 97.82 & 0.62 \\
w/o $DW_{1\times1}$ & 4.62 & 179 & 99.72 & 95.65 & 92.23 & 97.40 & 97.68 & 0.54 \\
\hline
\end{tabular}
\label{tab:repmix_abl}
\end{table}

\subsubsection{Ablation on Lite MHLA Configuration}
We conducted a series of ablation studies to examine the impact of different normalization strategies, activation functions, and token interaction mechanisms to further investigate the effectiveness of the Lite MHLA design, as summarized in Table~\ref{tab:lite_mhla_abl}. All experiments were performed using the FaceLiVTv2-S variant, trained on Glint360K to ensure a consistent comparison.

\begin{table}[t!]
\centering 
\setlength{\tabcolsep}{2.5pt} 
\caption{Lite MHLA Ablations of FaceLiVTv2-S}
\begin{tabular}{lcccccccc} 
\hline
\multirow{2}{*}{Ablations} & Param & FLOPs & \multirow{2}{*}{LFW} & CA &CP & CFP & Age & Lat \\ 
    & (M) & (M) &  & LFW & LFW & FP &DB30  & (ms) \\ \hline
\multicolumn{9}{c}{Normalization} \\
\hline
Layer Norm & 4.58 & 179 & 99.68 & 95.70 & 90.90 & 95.75 & 97.05 & 0.55 \\ 
\rowcolor{gray!20}
Affine & 4.62 & 179 & 99.78 & 95.93 & 92.45 & 97.47 & 97.82 & 0.54 \\
\hline
\multicolumn{9}{c}{Activation Function} \\
\hline
GELU & 4.62 & 179 & 99.76 & 95.90 & 92.38 & 97.44 & 97.79 & 0.55 \\
\rowcolor{gray!20}
None & 4.62 & 179 & 99.78 & 95.93 & 92.45 & 97.47 & 97.82 & 0.54 \\
\hline
\multicolumn{9}{c}{Token Interaction} \\
\hline
Additive Attn\cite{luevano2024swiftfaceformer} & 7.09 & 260 & 99.77 & 96.08 & 92.68 & 97.84 & 97.95 &  0.71 \\
\rowcolor{gray!20}
Lite MHLA & 4.62 & 179 & 99.78 & 95.93 & 92.45 & 97.47 & 97.82 & 0.54 \\
\hline
\end{tabular}
\label{tab:lite_mhla_abl}
\end{table}

\textit{Normalization}. Replacing the standard Layer Normalization (LN) with an affine transformation slightly improves overall accuracy while reducing latency. Specifically, the affine transformation variant achieves 99.77\% on LFW and 97.45\% on AgeDB-30 with a latency of 0.53~ms, compared to 99.72\% and 96.92\% obtained with LN. The affine transformation enhances numerical stability and feature calibration without introducing additional parameters, which is particularly beneficial in low-latency attention modules.

\textit{Non-linearity function}. We compared GELU against a linear formulation (no activation). Interestingly, removing the non-linearity leads to a small but consistent improvement across all test sets, suggesting that Lite MHLA benefits from a purely linear mapping in its token transformation stage. The absence of a non-linear activation likely reduces redundant feature distortion, allowing for smoother gradient flow and more stable optimization. Consequently, the linear variant achieves the highest performance, reaching 99.77\% on LFW and 97.45\% on AgeDB-30 while maintaining a minimal latency of 0.53~ms on the iPhone 15 Pro.

\textit{Token Interaction}. We evaluated linear projection against additive attention from SwiftFaceFormer~\cite{luevano2024swiftfaceformer, shaker2023swiftformer}. While additive attention achieves marginally higher accuracy on CP-LFW and CFP-FP, it requires 53.4\% more parameters, 45.3\% more FLOPs, and incurs 31.5\% slower inference than Lite MHLA. Critically, FaceLiVTv2-M with an equivalent computational budget outperforms additive attention on nearly all benchmarks, demonstrating that resources are better allocated to network scaling than complex token interactions. Lite MHLA's linear, activation-free formulation optimally balances accuracy and efficiency by reducing parameter redundancy and enhancing memory access patterns, which are critical for memory-bound mobile hardware.
\begin{table}[t!]
\centering 
\setlength{\tabcolsep}{2.75pt} 
\caption{Number of Head Ablations of Lite MHLA On FaceLiVTv2-S}
\begin{tabular}{ccccccccc} 
\hline
Num & Param & FLOPs & \multirow{2}{*}{LFW} & CA &CP & CFP & Age & Lat \\ 
Head & (M) & (M) &  & LFW & LFW & FP &DB30  & (ms) \\ \hline
1 & 4.55 & 179 & 99.70 & 95.87 & 91.88 & 97.20 & 97.75 & 0.49\\ \hdashline
\multirow{2}{*}{2} & \multirow{2}{*}{4.58} & \multirow{2}{*}{179} & 99.75 & 95.77 & 92.07 & 97.36 & 97.72 & 0.52\\
&      &     &\textbf{(+0.05)}&(-0.10)&\textbf{(+0.19)}&\textbf{(+0.16)}&(-0.03)& (6\%$\uparrow$)\\ \hdashline
\multirow{2}{*}{3} & \multirow{2}{*}{4.59} & \multirow{2}{*}{179} & 99.75 & 95.80 & 92.77 & 97.36 & 97.97 & 0.53\\
  &      &     &\textbf{(+0.05)}&(-0.07)&\textbf{(+0.89)}&\textbf{(+0.16)}&\textbf{(+0.22)}& (8\%$\uparrow$)\\ \hdashline
\rowcolor{gray!15}
  &  &  & 99.78 & 95.93 & 92.45 & 97.47 & 97.82 & 0.54\\ \rowcolor{gray!15}
\multirow{-2}{*}{4}&\multirow{-2}{*}{4.62}& \multirow{-2}{*}{179}&\textbf{(+0.08)}&\textbf{(+0.06)}&\textbf{(+0.57)}&\textbf{(+0.27)}&\textbf{(+0.07)}& (10\%$\uparrow$)\\ \hdashline
\multirow{2}{*}{5} & \multirow{2}{*}{4.64} & \multirow{2}{*}{179} & 99.75 & 95.75 & 92.90 & 97.33 & 97.85 & 0.55\\
  &      &     &\textbf{(+0.05)}&(-0.12)&\textbf{(+1.02)}&\textbf{(+0.13)}&\textbf{(+0.10)}& (12\%$\uparrow$)\\ \hdashline
\multirow{2}{*}{6} & \multirow{2}{*}{4.67} & \multirow{2}{*}{179} & 99.73 & 95.85 & 92.23 & 97.26 & 97.60 & 0.56\\
  &      &     &\textbf{(+0.03)}&(-0.02)&\textbf{(+0.35)}&\textbf{(+0.06)}&(-0.15)& (14\%$\uparrow$)\\ 
\hline
\end{tabular}
\label{tab:num_head_abl}
\end{table}

\subsubsection{Effect on the Number of Lite MHLA Heads} \label{sec:abl_num_hed}
We further investigate the effect of the number of Lite MHLA heads by varying the head count from one to six, using Glint360K as the training dataset. Table~\ref{tab:num_head_abl} reports the recognition accuracy together with the relative improvement over the single-head baseline. Increasing the number of heads consistently improves performance across most benchmarks, with particularly notable gains on CP-LFW and CFP-FP, indicating enhanced robustness to pose variation. When the number of heads is increased to three to five, accuracy improves on most benchmarks, while the inference latency increases only marginally from 0.49~ms to 0.55~ms. 

These results suggest that the multi-head formulation is more effective than a single linear token interaction, such as that adopted in ResMLP~\cite{touvron2022resmlp}, especially for challenging cross-pose and frontal–profile verification tasks. However, the performance gains saturate beyond five heads, with no further improvement observed.
This trend can be attributed to the reduction in per-head feature dimensionality as the number of heads increases, which may weaken discriminative capacity under a fixed embedding size. In our experiments, four heads provide a favorable balance between interaction diversity, per-head expressiveness, recognition accuracy, and computational efficiency, while additional heads introduce redundancy without yielding further performance gains.
\begin{figure}[t!]
    \centering
    \includegraphics[width=\columnwidth]{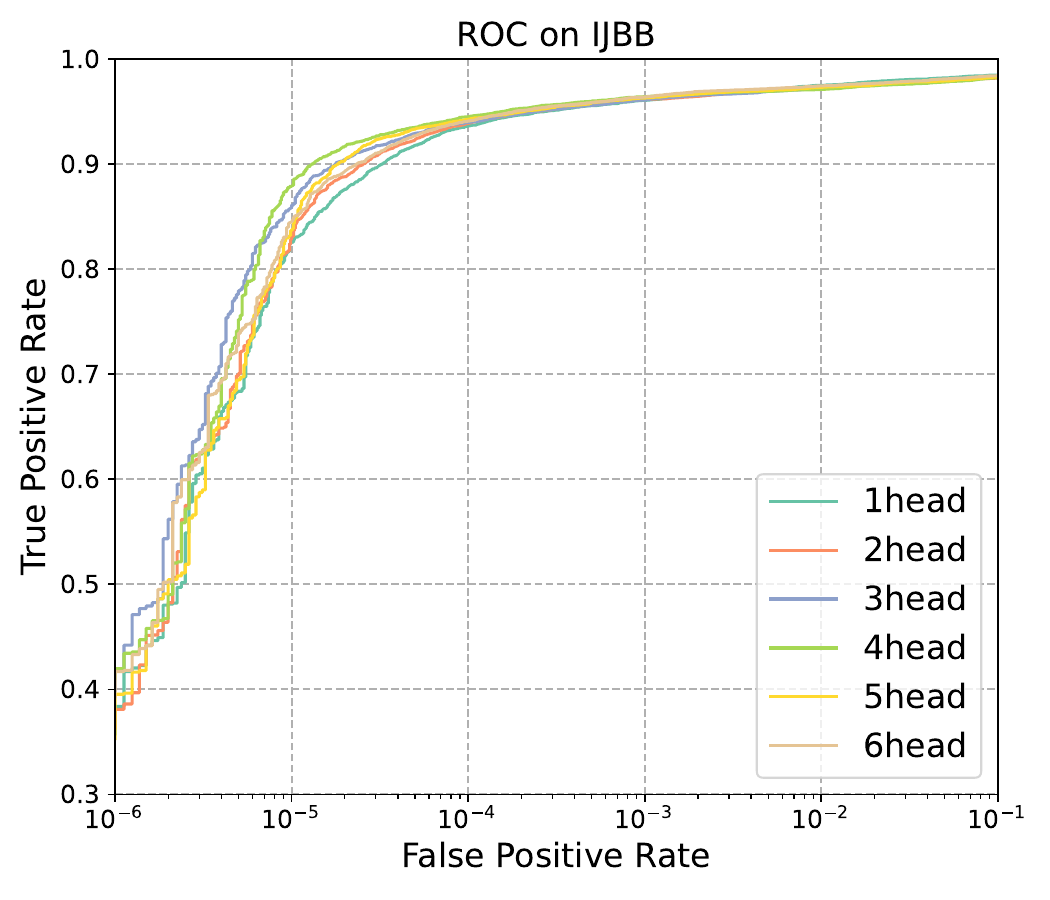}
    \includegraphics[width=\columnwidth]{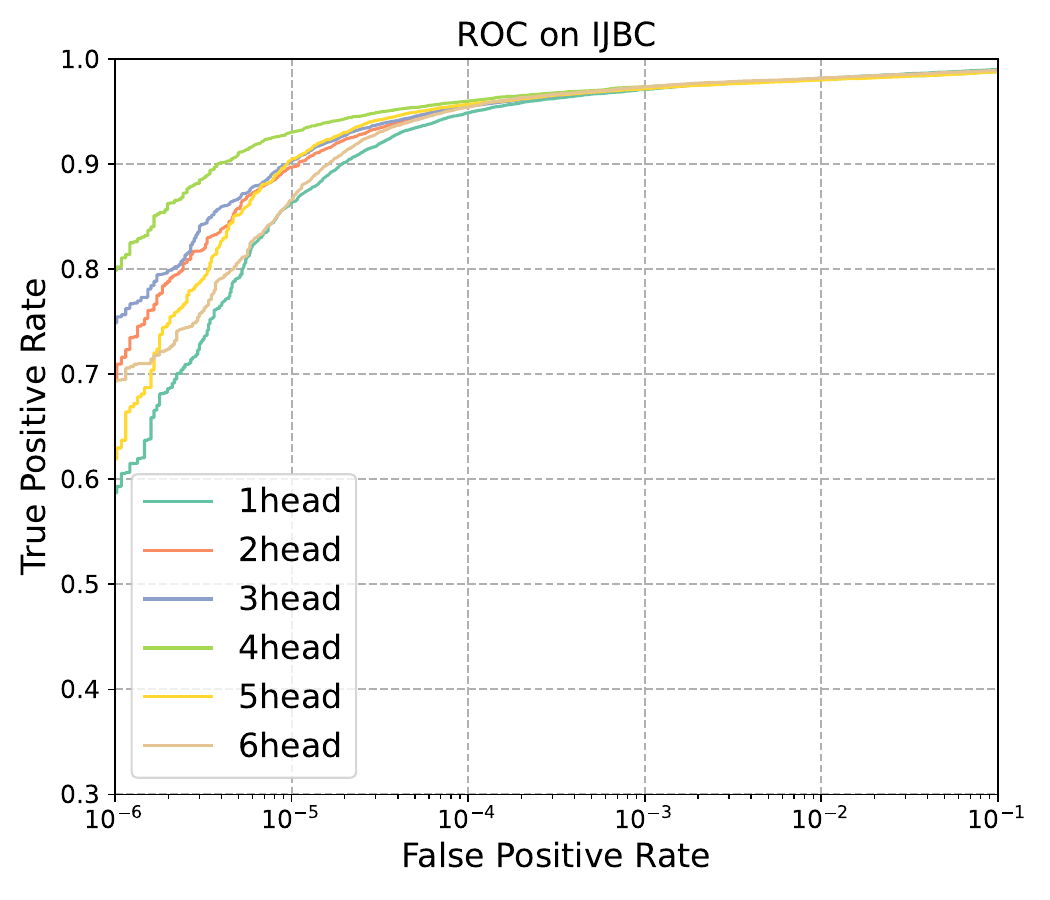}
    \caption{Performance evolution for FaceLiVTv2-S across different number of head configuration on IJBB and IJBC Dataset}
    \label{fig:ijbb_roc}
\end{figure} 

\begin{table}[t!]
\centering
\setlength{\tabcolsep}{3.0pt}
\caption{Performance evolution of FaceLiVTv2-S with different numbers of Lite MHLA heads on the IJB-B and IJB-C Test}
\begin{tabular}{cllll}
\hline
Num  & \multicolumn{2}{c}{IJB-B} & \multicolumn{2}{c}{IJB-C} \\ \cline{2-5}
Head & FAR@$10^{-6}$ & FAR@$10^{-4}$ & FAR@$10^{-6}$ & FAR@$10^{-4}$ \\ \hline
1 & 38.38 & 93.62 & 59.31 & 94.89 \\ 
2 & 38.10\,(-0.28) & 93.89\,(+0.27) & 70.96\,(+11.65) & 95.50\,(+0.61) \\ 
3 & 41.76\,(+3.38) & 93.96\,(+0.34) & 75.45\,(+16.14) & 95.56\,(+0.67) \\ \rowcolor{gray!15}
4 & \textbf{41.97\,(+3.59)} & \textbf{94.51\,(+0.89)} & \textbf{80.22\,(+20.91)} & \textbf{95.99\,(+1.10)} \\ 
5 & 39.53\,(+1.15) & 94.26\,(+0.64) & 62.97\,(+3.66) & 95.67\,(+0.78) \\ 
6 & 41.69\,(+3.31) & 94.10\,(+0.48) & 69.40\,(+10.09) & 95.40\,(+0.51) \\ 
\hline
\end{tabular}
\label{tab:num_head_ijb}
\end{table}

The verification and identification performance on the IJB-B and IJB-C datasets further validate these findings, as shown in Table~\ref{tab:num_head_ijb}. Increasing the number of heads enhances the model’s ability to capture discriminative global context, improving the true-accept rate (TAR) at very low false-accept rates (FAR). The four-head configuration achieves the best verification accuracy on IJB-B with 41.97\% TAR@${10^{-6}}$ and 94.51\% TAR@${10^{-4}}$, while on IJB-C it obtains 80.22\% TAR@${10^{-6}}$ and 95.99\% TAR@${10^{-4}}$. As illustrated in Fig.~\ref{fig:ijbb_roc}, the ROC curves confirm that the four-head configurations deliver the highest operating-point performance with minimal overhead. This test shows the advantages of the designed multi-head model in achieving performance robustness under real-world and unconstrained conditions.

Overall, these results demonstrate that Lite MHLA benefits from multiple heads up to an optimal point, where additional heads no longer improve accuracy but linearly increase computation. Consequently, the four-head Lite MHLA configuration is adopted as the default setting in FaceLiVTv2, providing the best trade-off between global modeling capability, computational efficiency, and verification robustness on large-scale IJB benchmarks.

\subsection{Limitations and Future Work}
While FaceLiVTv2 demonstrates strong generalization across diverse benchmarks—including CA-LFW, CP-LFW, AgeDB-30, IJB-B/C, and the low-resolution TinyFace dataset—several limitations remain. The current study does not explicitly assess robustness under extreme occlusions, such as heavy facial masking or severe truncation, nor does it analyze cross-ethnicity generalization and demographic bias due to the lack of fine-grained race and gender annotations in the employed benchmarks.

Future work will address these gaps by incorporating occlusion-focused benchmarks and demographically annotated datasets for systematic fairness evaluation. We also plan to explore diffusion-based face restoration combined with knowledge distillation to build a unified lightweight pipeline for low-resolution FR on mobile devices, motivated by our TinyFace results suggesting that explicitly recovering fine-grained facial details through generative priors could further complement Lite MHLA's global modeling capacity.

\section{Conclusion}

In this paper, we present FaceLiVTv2, an enhanced lightweight transformer-based architecture for efficient face recognition on limited-resource devices, especially mobile platforms. Building upon the original FaceLiVT, the proposed model co-designs three key improvements: a Lite MHLA module for efficient global dependency modeling, a GDConv-based adaptive spatial aggregation head, and a unified RepMix-Lite MHLA structure that reinforces local-global feature interaction while maintaining computational efficiency. Extensive evaluations across eight benchmarks, including LFW, AgeDB-30, CFP-FP, IJB-B, IJB-C, and the low-resolution TinyFace, confirm that FaceLiVTv2 consistently outperforms both its predecessor and recent lightweight alternatives, achieving higher accuracy with fewer parameters and lower on-device latency. We hope that FaceLiVTv2 serves as a useful reference for designing efficient and deployable face recognition systems on resource-constrained platforms.

\section{Acknowledgment}
This research was supported by the National Science and Technology Council, Taiwan, through Grant Number NSTC-113-2221-E-305-018-MY3. The authors are grateful for the support in this research.

\bibliography{ref}
\bibliographystyle{ieeetr}

\begin{IEEEbiography}[{\includegraphics[width=1in,height=1.25in,clip,keepaspectratio]{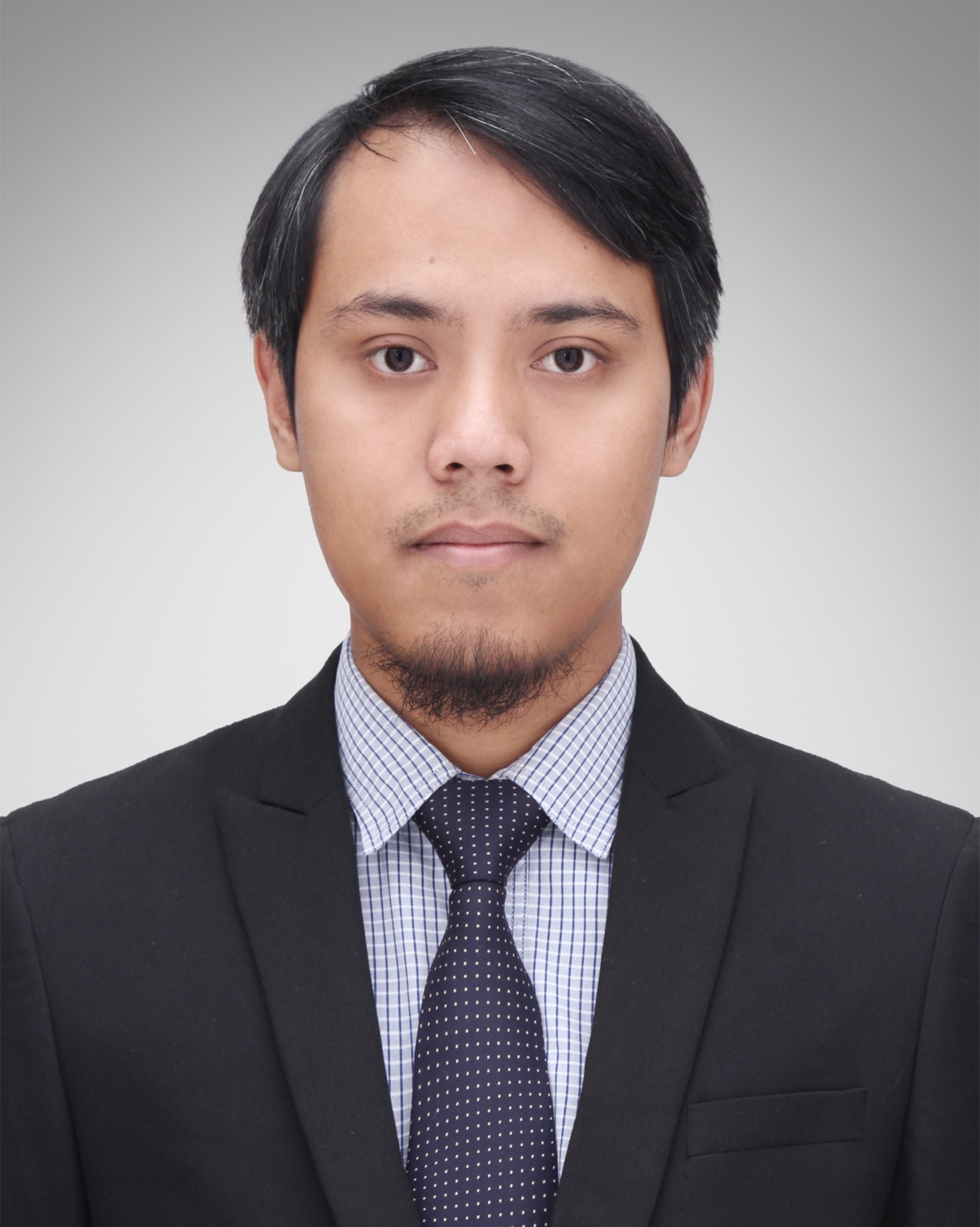}}]{Novendra Setyawan (Student Member, IEEE)}{\space}received the B.Eng. degree in Electrical Engineering from the University of Muhammadiyah Malang, Indonesia, in 2015, and the M.Eng. degree in Electronic Engineering from the Institut Teknologi Sepuluh Nopember (ITS), Surabaya, Indonesia, in 2017. He is currently pursuing the Ph.D. degree in the Department of Electro-Optics Engineering, National Formosa University, Taiwan. He is also a Lecturer in the Department of Electrical Engineering, University of Muhammadiyah Malang, Indonesia. His research has been published in IEEE APCCAS, IEEE ISCAS, IEEE ICIP, and other international venues. His research interests include deep learning based image processing, computer vision, and efficient model design for resource-constrained platforms.
\end{IEEEbiography}
\vskip -1.5\baselineskip plus -1fil
\begin{IEEEbiography}[{\includegraphics[width=1in,height=1.25in,clip,keepaspectratio]{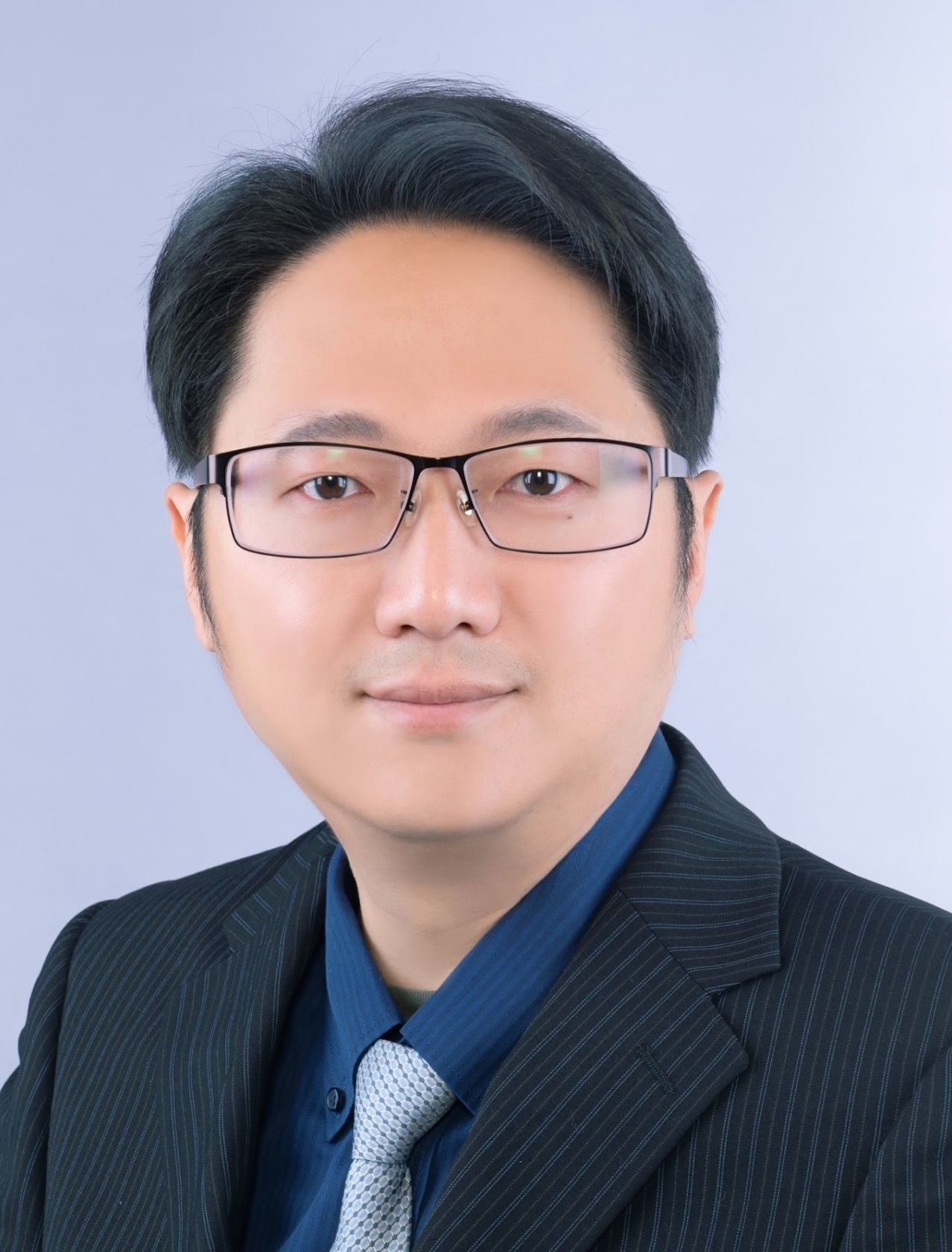}}]{Chi-Chia Sun (Member, IEEE)}{\space} received the B.S. degree in computer science and engineering from National Taiwan Ocean University, Taiwan, in 2004, the M.S. degree in electronic engineering from the National Taiwan University of Science and Technology, Taipei, Taiwan, in 2006, and the Dr.-Ing. degree (with Federal Republic of Germany DAAD Full Scholarship) from Dortmund University of Technology, Dortmund, Germany, in 2011. He was a Principle Engineer with TSMC, specializing in standard cell design. He is currently a Full Professor with the Department of Electrical Engineering, National Taipei University, Taipei. His research interests include image processing, system integration, and VLSI/FPGA design. He is a member of IEEE CASS VSA-TC.
\end{IEEEbiography}
\vskip -1.5\baselineskip plus -1fil
\begin{IEEEbiography}
[{\includegraphics[width=1in,height=1.25in,clip,keepaspectratio]{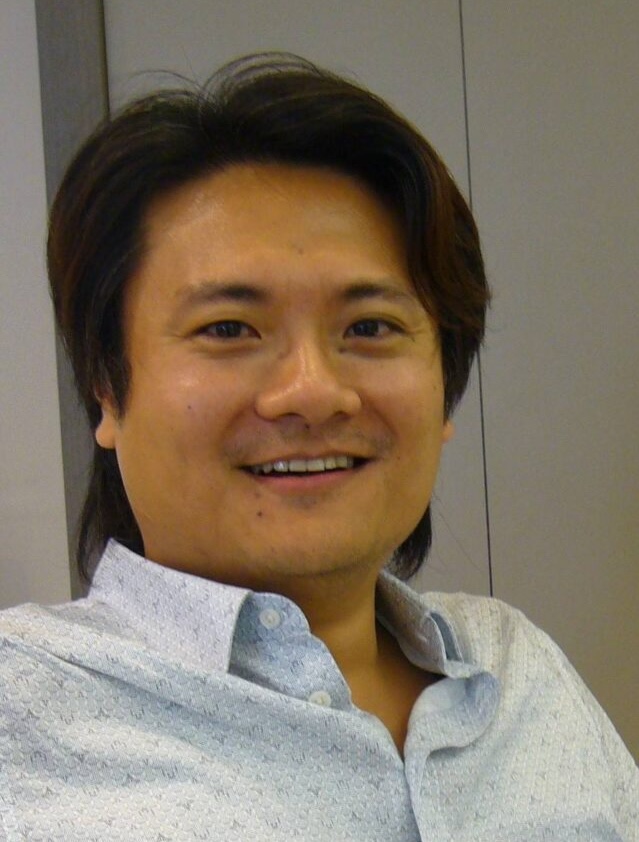}}]{Mao-Hsiu Hsu (Member, IEEE)}{\space} obtained his Ph.D. in electronic and computer engineering from National Taiwan University of Science and Technology in 2005. His career began in 1997 at Motorola, working on crystal filter and oscillator design. He joined Foxconn in California from 2007 to 2012, and in 2013, he was the Inpaq USA Site Country Manager. From 2014 to 2019, he was a Senior Manager at Primax Taiwan. He is currently an Associate Professor in Electro-Optical Engineering at National Formosa University, having previously served as an Assistant Professor in the same department from 2023 to 2025. He possesses extensive expertise in sensor and optical signal processing and deep learning algorithms for fingerprint and facial recognition sensor IC systems, with over 100 technical publications and patents. His current research focuses on sensing systems for Diffusion face ID, fingerprint detection under DeepFake GAN machine learning, and high-speed SerDes re-timer and Tiny ML IC design.   
\end{IEEEbiography}
\vskip -1.5\baselineskip plus -1fil
\begin{IEEEbiography}[{\includegraphics[width=1in,height=1.25in,clip,keepaspectratio]{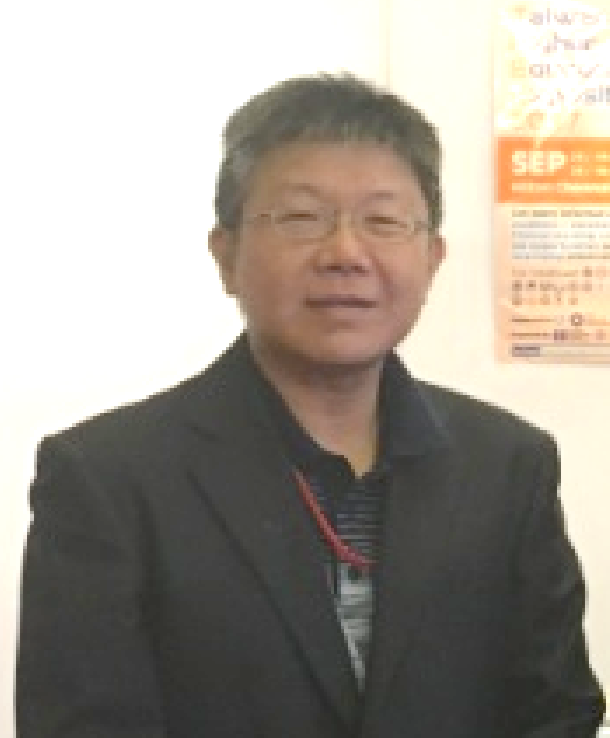}}]{Wen-Kai Kuo (Member, IEEE)}{\space} received the Ph.D. degree in electronic engineering from National Chiao Tung University, Hsin-Chu, Taiwan, in 2000. He is currently a Professor in the Department of Electro-Optics Engineering, National Formosa University, Huwei, Yunlin, Taiwan, where he has been a faculty member since 2000. He is a member of the Phi-Tau-Phi Honorary Scholar Society. His research interests include optical sensors, optical signal processing and analysis with artificial intelligence approaches, and optical systems. He has been involved in multiple research projects supported by the National Science and Technology Council (NSTC), Taiwan, with contributions spanning optical measurement systems and AI-based optical analysis.
\end{IEEEbiography}
\vskip -1.5\baselineskip plus -1fil
\begin{IEEEbiography}[{\includegraphics[width=1in,height=1.25in, clip,keepaspectratio]{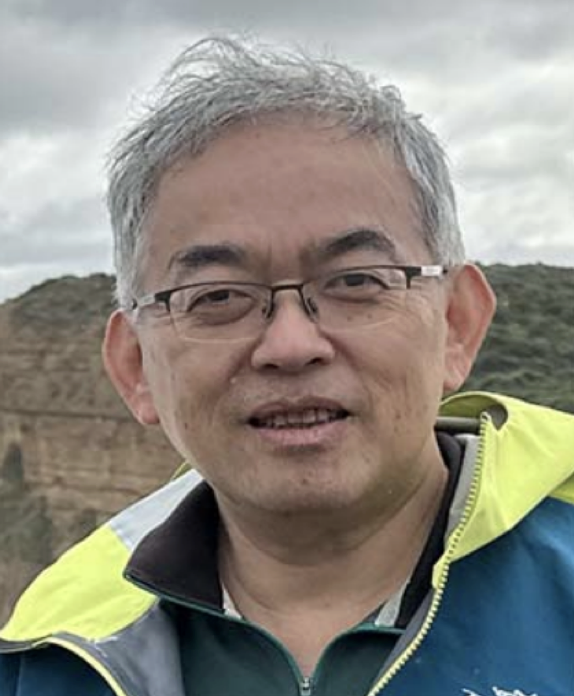}}]{Jun-Wei Hsieh (Senior Member, IEEE)}{\space} received the Ph.D. degree in computer engineering from the National Central University, Chung-Li, Taiwan, in 1995. He was an Associate Professor with the Department of Electrical Engineering, YuanZe University, and a Visiting Researcher with the MIT AI Laboratory. Since August 2009, he had been a Professor and the Dean of the Department of Computer Engineering, National Taiwan Ocean University. After August 2019, he has been a Professor with the College of AI, National Yang-Ming Chiao-Tung University. He hosted or co-hosted a lot of large-scale AI projects from different companies and governments in the past. He has a lot of successful experiences in industrial-academic cooperation and technology transferring, especially in ITS. His research interests include AI, deep learning, smart farming, video surveillance, intelligent transportation systems, image and video processing, object recognition, machine learning, 3D printing, medical image analysis, and computer vision. In May 2019, he received the First Prize of the Ministry of Science and Technology Best Display Award and the Third Place of the AI Investment Potential Award. Due to his contributions in traffic flow estimation, he helped the Elan company received the Gold Award from Taipei International Computer Show, in 2019. He and his students received the Silver Medal of 2019 National College Software Creation Competition, the Silver Medal of 2018 National Microcomputer Competition, the Best Paper awards of Information Technology and Applications in Outlying Islands Conference, in 2013, 2014, 2016, 2017, 2018, 2021, and 2022, respectively, the Best Paper Award of Tanet 2017, the Best Paper Awards of NCWIA 2020, 2021, and 2022, respectively, and the Best Paper Awards of IS3C 2020.
 He also received the Best Paper Award of CVGIP Conference, in 1999, 2003, 2005, 2007, 2014, 2017-2018, and 2022-2024, the Best Paper Award of DMS Conference, in 2011, the Best Paper Award of IIHMSP 2010, and the Best Patent Award of Institute of Industrial Technology Research, in 2009 and 2010, respectively. Dr. Hsieh serves as Program Chair of Conference on Multimedia Modeling 2011, and Program Chair of the IEEE Advanced Video and Signal-based Surveillance (AVSS) 2019 and 2025. He has authored more than 150 peer-reviewed journal and conference publications, and 20 US/Taiwan patents.
\end{IEEEbiography}
\break 
\begin{IEEEbiographynophoto}{\relax}\unskip
\end{IEEEbiographynophoto}

\twocolumn[
\vspace{3.0em}
\centering
\Large
\bigskip
\bigskip
\textbf{FaceLiVTv2: An Improved Hybrid Architecture for Efficient Mobile Face Recognition }

\textbf{------ Supplementary Material ------} \\
\vspace{3.0em}

]

In this supplementary material, we present implementation details and additional quality comparisons to further demonstrate the effectiveness and quality of our proposed method in comparison to other existing approaches. By including these extended evaluations, we aim to provide a more comprehensive analysis and highlight the advantages of our method over alternative techniques.
\section*{Implementation Details of FaceLiVTv2 }
The advantages of FaceLiVTv2 are using RepMixer, which can be reparameterized while inferring the model to speed up. Below is the implementation of RepMixer in PyTorch-like code:
  \begin{lstlisting}[language=Python]
class RepMix(nn.Module):
  def __init__(self, c1, c2, ks=1, s=1, p=0, g=1):
    super().__init__()
    self.cv = nn.Conv2d(c1, c2, ks, s, p, groups=g)
    p = p//2; ks = ks//2
    self.rcv = nn.Conv2d(c1, c2, ks, s, p, groups=g)
    self.bn = nn.BatchNorm2d(ouc)
    
  def forward(self, x):
    xr = self.cv(x) + self.rcv(x) 
    return self.bn(xr)
    
  def forward_deploy(self, x):
    return self.cv(x)
    
  @torch.no_grad()
  def reparam(self):
    cv = self.cv
    rcv=self.rcv; self.__delattr__('rcv')
    kw = (cv.weight.shape[2]-rcv.weight.shape[2])//2 
    kh = (cv.weight.shape[3]-rcv.weight.shape[3])//2
    rcv_w = F.pad(rcv.weight, [kh,kh,kw,kw])
    rcv_b = rcv.bias         
    final_cv_w = cv.weight + rcv_w 
    final_cv_b = cv.bias + rcv_b 
    cv.weight.data.copy_(final_cv_w)
    cv.bias.data.copy_(final_cv_b)
    bn = self.bn
    w = bn.weight / (bn.running_var + bn.eps)**0.5
    w = cv.weight * w[:, None, None, None]
    b = bn.bias + (cv.bias - bn.running_mean) * 
    bn.weight / (bn.running_var + bn.eps)**0.5
    cv.weight.data.copy_(w)
    cv.bias.data.copy_(b)
    self.__delattr__('bn')
    self.forward = self.forward_deploy
    self.cv = cv
    return self
\end{lstlisting}
The other advantages of FaceLiVTv2 are using Lite Multi-Head Linear Attention (Lite MHLA), which is an improved version of the previous MHLA in FaceLiVTv1. Below is the implementation of Lite MHLA in PyTorch-like code:

\begin{lstlisting}[language=Python]
class LiteMHLA(torch.nn.Module):
  def __init__(self, dim, resolution):
    super().__init__()
    self.n_head = 4
    self.res=resolution**2
    self.dim=dim
    linear=[]

    self.norm = Affine(dim) 
    for i in range(self.n_head):
        lin = nn.Linear(self.res, self.res)
        linear.append(lin)
    self.lin = torch.nn.ModuleList(linear)
    ls_init = torch.ones(dim).unsqueeze(-1)
    ls_init = ls_init.unsqueeze(-1)
    self.ls = nn.Parameter(1e-5 * ls_init)

  def forward(self, x):
    B,C,H,W = x.shape
    x = x.reshape(-1, self.dim, self.res)
    x = self.norm(x)
    x = list(x.chunk(self.n_head, dim=1))
    for i in range(self.n_head):
        x[i] = self.lin[i](x[i])
    x = torch.cat(x, dim=1)
    x =  self.ls * x.reshape(B,C,H,W)
    return x
\end{lstlisting}

We employ a similar training methodology as \cite{george2024edgeface, martinez2021benchmarking, martindez2019shufflefacenet} on Glint360K \cite{an2021partial}. The AdamW optimizer and a polynomial decay learning rate schedule are utilized. The starting learning rate is $6\times10^{-3}$ with a minimum of $1\times10^{-5}$. The batch size is 342 on each $3\times$ NVIDIA RTX-A6000 GPU, and the weight decay is $1\times10^{-4}$. The model is trained for 50 epochs with a pre-processing resolution of $(112\times 112)$. 
 
    \begin{table}[htb]
        \centering
        \caption{FaceLiVT Training Hyperparameters and Augmentation on Glint360K datasets}
        \begin{tabular}{c|c}
        \hline
        Hyperparameters & Config \\ \hline
        optimizer & AdamW \\
        learning rate & 0.006 \\
        image resolution & 112$\times$112 \\
        batch size &  384 \\
        LR schedule & polynomial \\
        warmup epochs & 0 \\ 
        training epochs & 50 \\
        weight decay & 0.01 \\
        embedding size & 512 \\
        loss function & CosFace \\
        loss hyperparam & [1, 0.4] \\
        random flip     & 0.5 \\
        normalization mean & [0.5, 0.5, 0.5] \\
        normalization std & [0.5, 0.5, 0.5] \\
        \hline
        \end{tabular}
        \label{tab:hyper}
    \end{table}

\section*{Detailed Result On IJB-B and IJB-C robustness Test}

Table~\ref{tab:ijbbenchmark} and Fig.~\ref{fig:ijbroc} present a comprehensive comparison of the proposed FaceLiVTv2 variants with state-of-the-art lightweight and standard face recognition models on the IJB-B and IJB-C benchmark datasets. The results clearly demonstrate the effectiveness and scalability of the proposed architecture across model sizes (XS $\rightarrow$ L).

\begin{table}[!t]
\centering
\setlength{\tabcolsep}{2.5pt} 
\renewcommand{\arraystretch}{1.1}
\caption{Comparison of FaceLiVTv2 Variant with State-Of-The-Art on IJB-B and IJB-C Benchmark Dataset.}
\begin{tabular}{ l|ccc|ccc } 
\hline
\multirow{2}{*}{Model} & \multicolumn{3}{c|}{IJB-B}& \multicolumn{3}{c}{IJB-C} \\ \cline{2-7} 

& $1e^{-5}$& $1e^{-4}$ & $1e^{-3}$ & $1e^{-5}$& $1e^{-4}$ & $1e^{-3}$   \\ \hline
MobileFaceNet\cite{martinez2021benchmarking}  & 87.9 & 92.8 & 95.6 & 92.2 & 94.7 & 96.6\\
ShuffleFaceNet-1.5\cite{martinez2021benchmarking} & 86.5 & 92.3 & 95.2 & 91.3 & 94.3 & 96.3\\ 
EdgeFace-S(0.5) \cite{george2024edgeface} & - & 93.6 & - & - & 95.6 & - \\
ResNet50-ArcFace \cite{deng2019arcface}  & 80.5 & 89.9 & 94.5 & 86.1 & 92.1 & 96.0 \\
EdgeFace-XS(0.6) \cite{george2024edgeface} & - & 92.7 & - & - & 94.8 & - \\ 

\hline
FaceLiVT-S    & - & 91.2 & - & - & 92.7 & - \\
FaceLiVT-M    & - & 93.7 & - & - & 95.7 & - \\
FaceLiVTv2-XS & 72.38 & 90.67 & 95.38 & 75.27 & 91.25 & 96.29 \\
FaceLiVTv2-S  & 87.94 & 94.51 & 96.41 & 93.04 & 95.99 & 97.37 \\
FaceLiVTv2-M  & 88.92 & 95.02 & 96.59 & 93.59 & 96.42 & 97.62 \\
FaceLiVTv2-L  & 91.09 & 95.18 & 96.57 & 94.52 & 96.59 & 97.55 \\
\hline
\end{tabular}
\label{tab:ijbbenchmark}
\end{table}

\section*{Comparison with Existing Lightweight Models}
Compared with MobileFaceNet, ShuffleFaceNet-1.5, and EdgeFace variants, all FaceLiVTv2 models achieve substantially higher true positive rates, especially at stringent false positive rates of $10^{-5}$ and $10^{-4}$. 
For instance, FaceLiVTv2-S reaches 94.51\% on IJB-B and 95.99\% on IJB-C at $10^{-4}$, outperforming ShuffleFaceNet-1.5 by +2.2\% and +1.6\%, respectively. 
This performance gain highlights the advantage of the proposed \textit{Lite Multi-Head Linear Attention (Lite-MHLA)} and \textit{Global Depthwise Convolution (GDConv)} modules in preserving discriminative features under low false acceptance thresholds.

\section*{Scaling Behavior within FaceLiVTv2 Family}
The ROC curves in Fig.~\ref{fig:ijbroc} show a consistent upward trend from \textit{FaceLiVTv2-XS} to \textit{FaceLiVTv2-L}, indicating that model scaling directly enhances representation power without sacrificing inference stability. 
The XS variant already surpasses most existing lightweight baselines, while the \textit{L model} achieves near-saturation accuracy of 95.18\% (IJB-B) and 96.59\% (IJB-C) at $10^{-4}$. 
The S and M variants provide the best trade-off between accuracy and efficiency, making them suitable for deployment on mobile or edge devices.

\section*{ROC Curve Insights}
\begin{figure}[t!]
    \centering
    \includegraphics[width=\columnwidth]{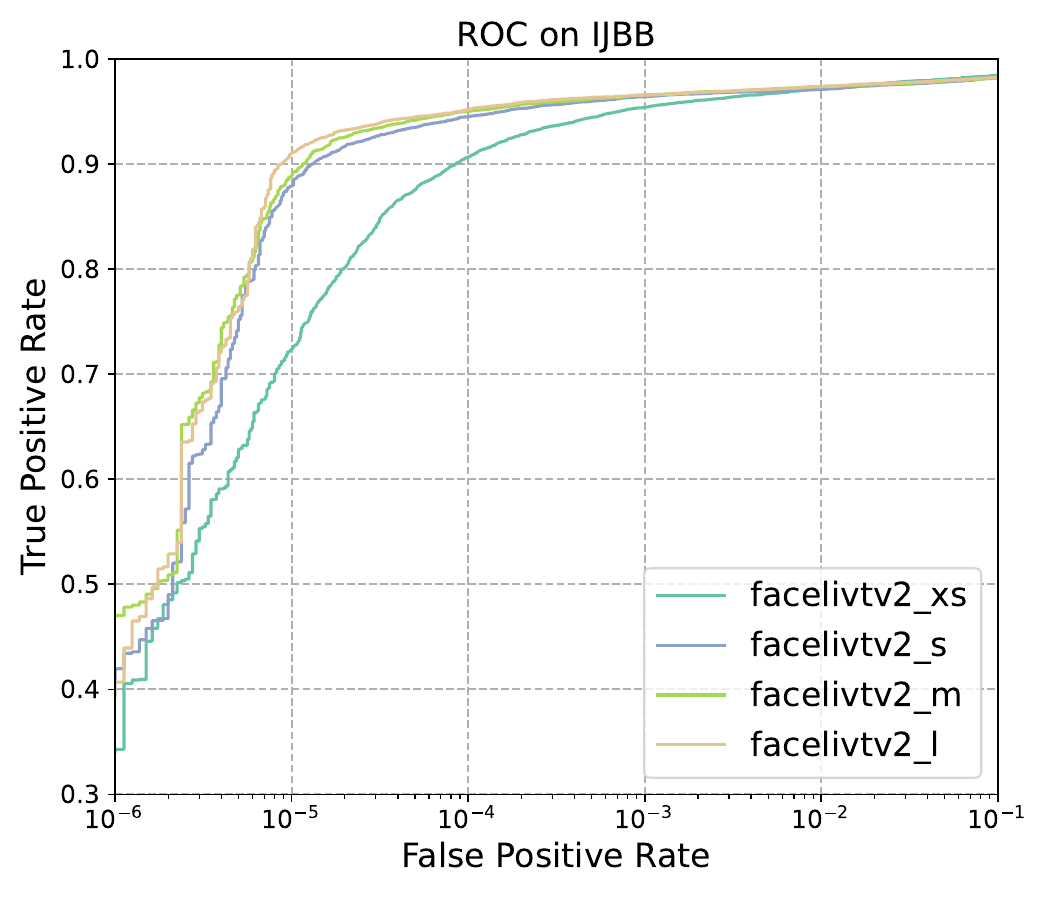}
    \includegraphics[width=\columnwidth]{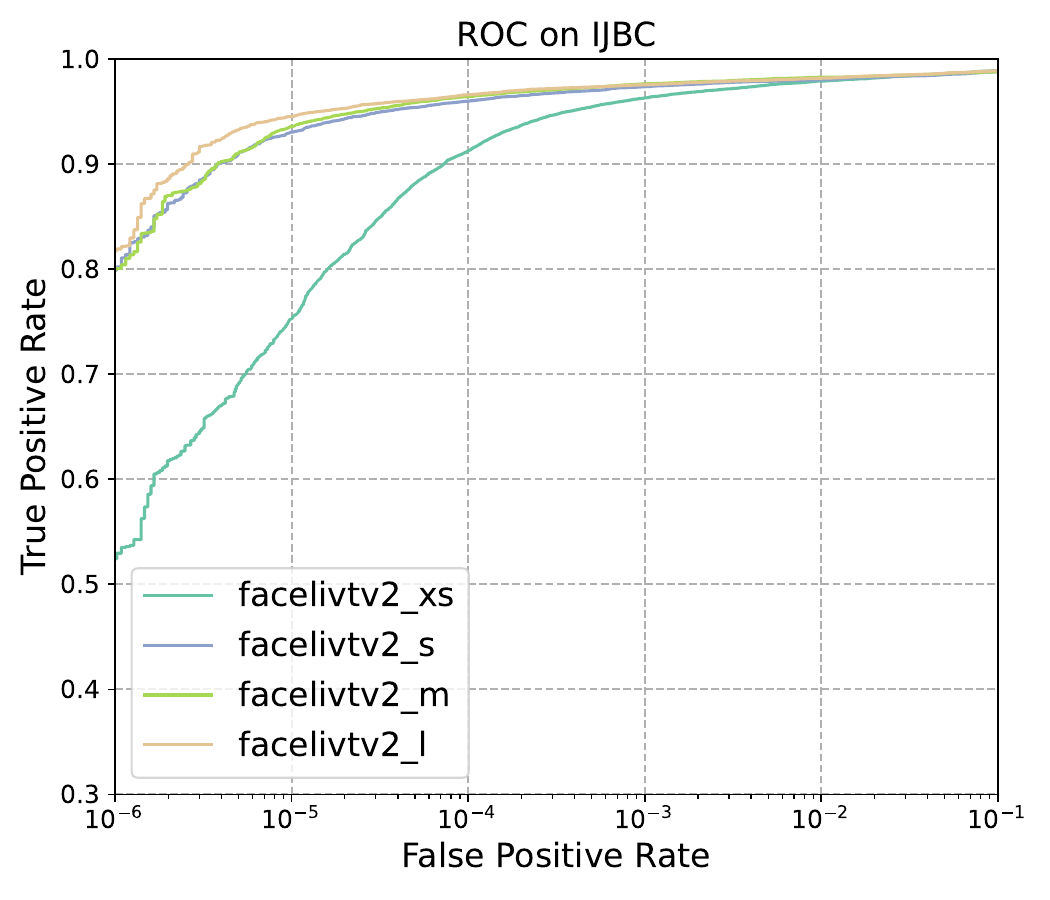}
    \caption{Receiver Operating Characteristic of FaceLiVTv2 on IJBB and IJBC Dataset}
    \label{fig:ijbroc}
\end{figure}

As depicted in Fig.~\ref{fig:ijbroc}, the ROC curves of \textit{FaceLiVTv2} models consistently lie above those of previous \textit{FaceLiVT} and baseline networks. 
The sharper rise in true positive rate at very low false positive regions ($10^{-6} \leq \text{FPR} \leq 10^{-4}$) confirms that the attention reparameterization and spatial aggregation modules effectively enhance robustness to pose and illumination variations. 
The performance gap between XS and S variants narrows at higher FPR values, suggesting diminishing returns for larger models once feature discriminability reaches saturation.

Overall, \textit{FaceLiVTv2} demonstrates clear superiority over CNN-based approaches (MobileFaceNet, ShuffleFaceNet) and hybrid (EdgeFace) baselines, achieving accuracy gains of up to +4\% at $10^{-5}$ on IJB-B. 
These results validate the design goal of achieving both compactness and high accuracy, making \textit{FaceLiVTv2} highly adaptable for real-time and low-power applications on mobile and edge platforms.

\section*{Latency Evaluation Setup}
The inference latency of all FaceLiVTv2 variants was measured using an iPhone 15 Pro (A17 Pro chip, 128-GB) running iOS-26.0.1. Each model was first trained in the PyTorch framework, exported to ONNX, and then converted to Apple’s CoreML format using the \texttt{coremltools} library. The conversion was performed with the \texttt{mlmodel} specification and executed on all available compute units to ensure compatibility with Apple’s Neural Engine. The following Python command was used to perform the conversion:
\begin{lstlisting}[language=Python, frame=None] 
    import torch, coremltools as ct

    inp = torch.rand(1, 3, res, res) 
    traced_model = torch.jit.trace(model, 
                   torch.Tensor(inp))
    out = traced_model(inp)
    model = ct.convert(traced_model,
            convert_to="neuralnetwork",
            inputs=[ct.ImageType(shape=inp.shape)])
 
    model.save(f"./{args.model}_{res}.mlmodel")
\end{lstlisting}

The converted Core~ML models were deployed on-device using a lightweight iOS inference application developed in \texttt{Xcode-16.0}. Each model received a normalized 112$\times$112 RGB input tensor, and the inference latency was measured using the \texttt{Core~ML Performance Template} available in \texttt{Xcode Instruments}. The profiling environment was configured to execute exclusively on the Neural Engine to evaluate on-device hardware acceleration performance. The profiler automatically reported three latency components: (1)~prediction time (per-image inference duration), (2)~model load time (initialization and memory allocation overhead), and (3)~compilation time (graph optimization and hardware binding). 

\begin{figure} [!t]
    \centering
    \includegraphics[width=9cm]{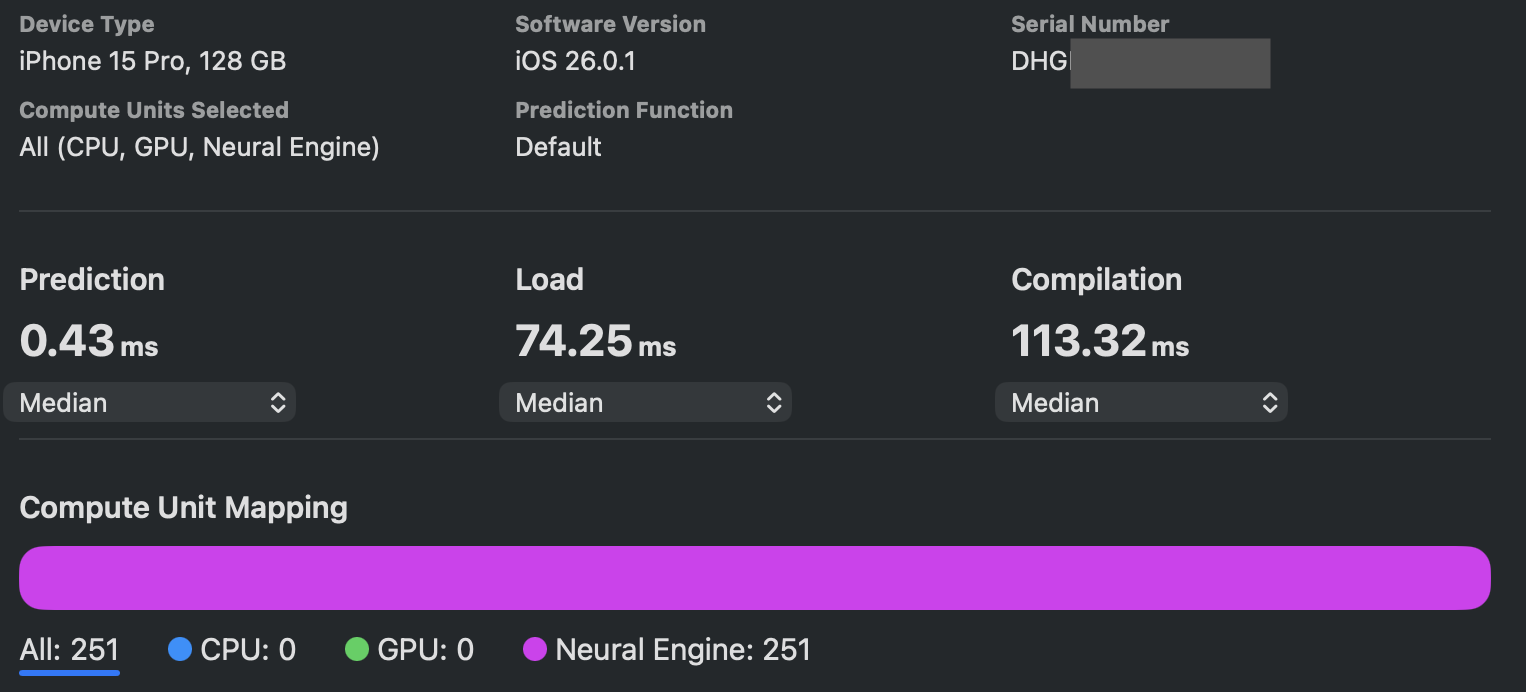}
    \\(1)\hspace{8cm} \\
    \includegraphics[width=9cm]{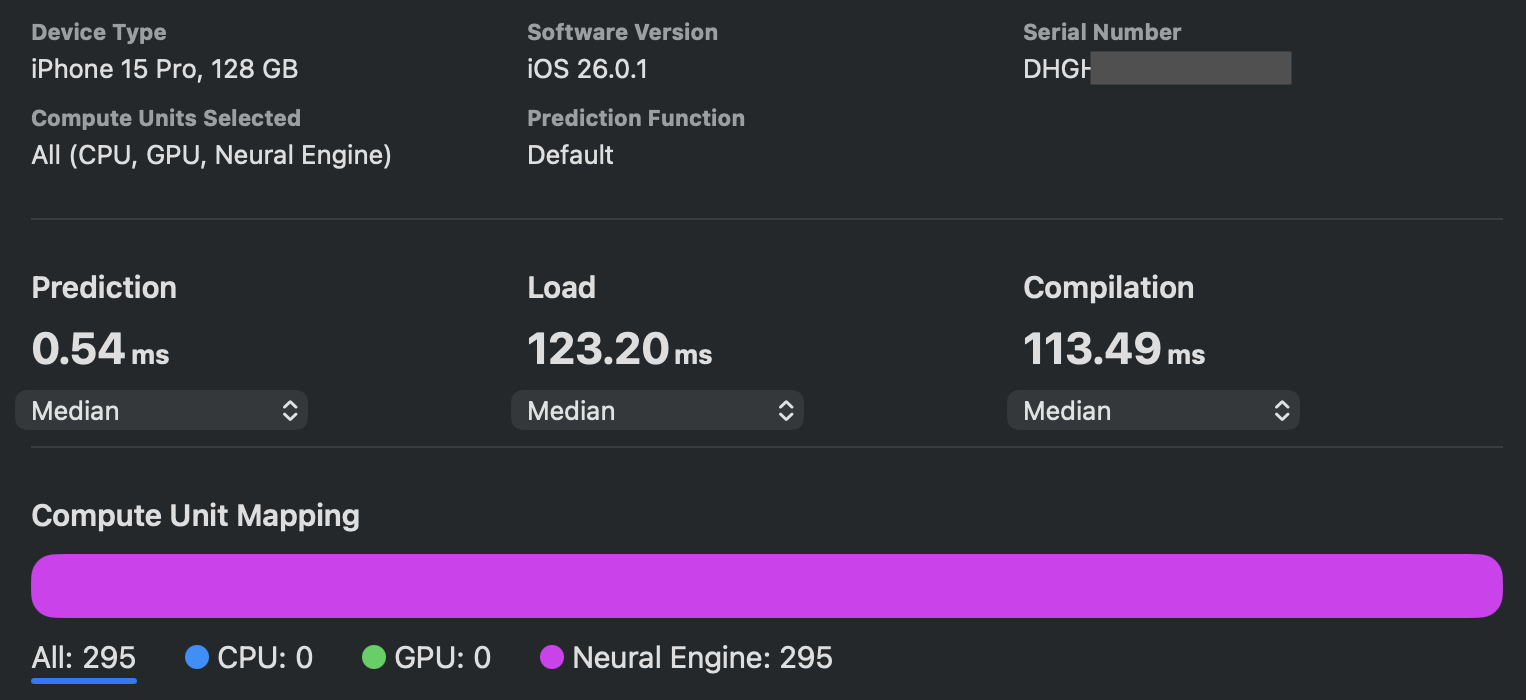}
    \\(2)\hspace{8cm} \\ 
    \includegraphics[width=9cm]{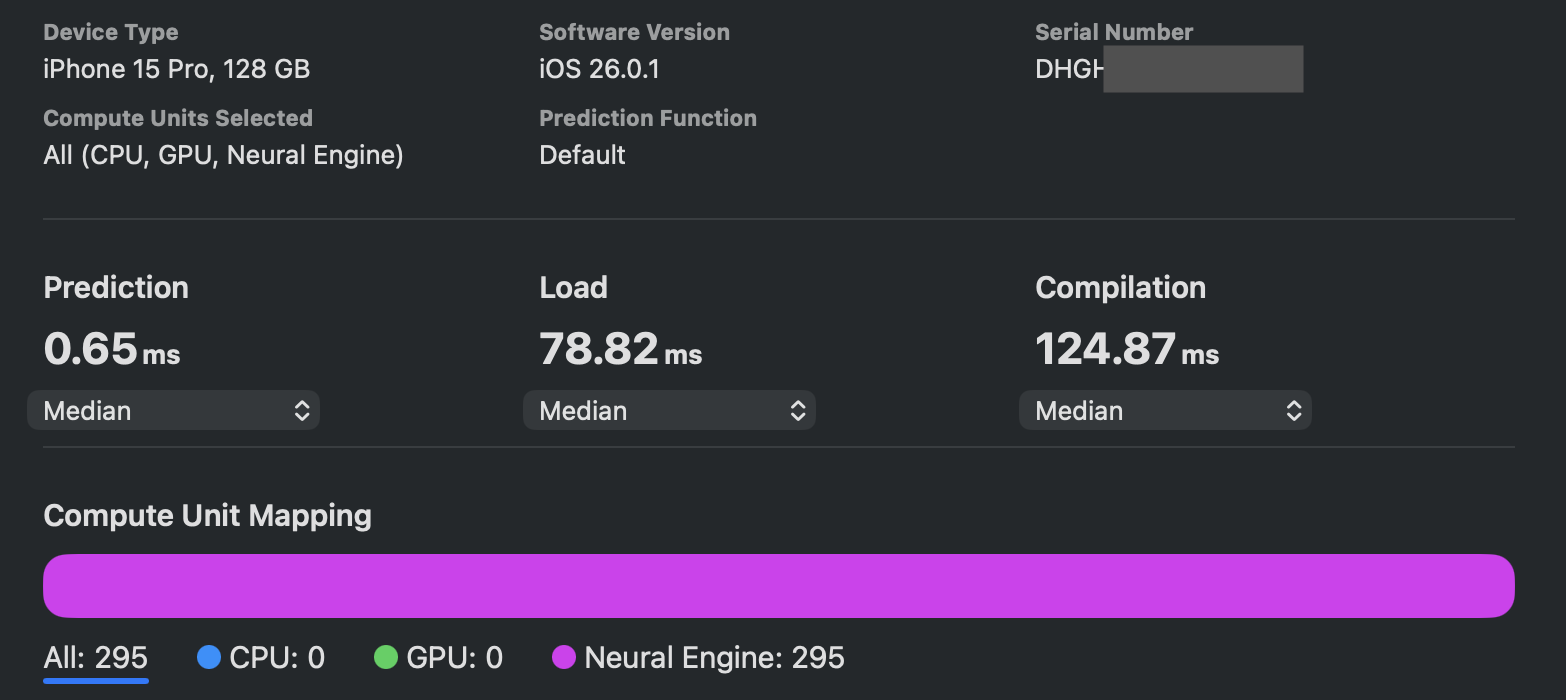}
    \\(3)\hspace{8cm} \\
    \includegraphics[width=9cm]{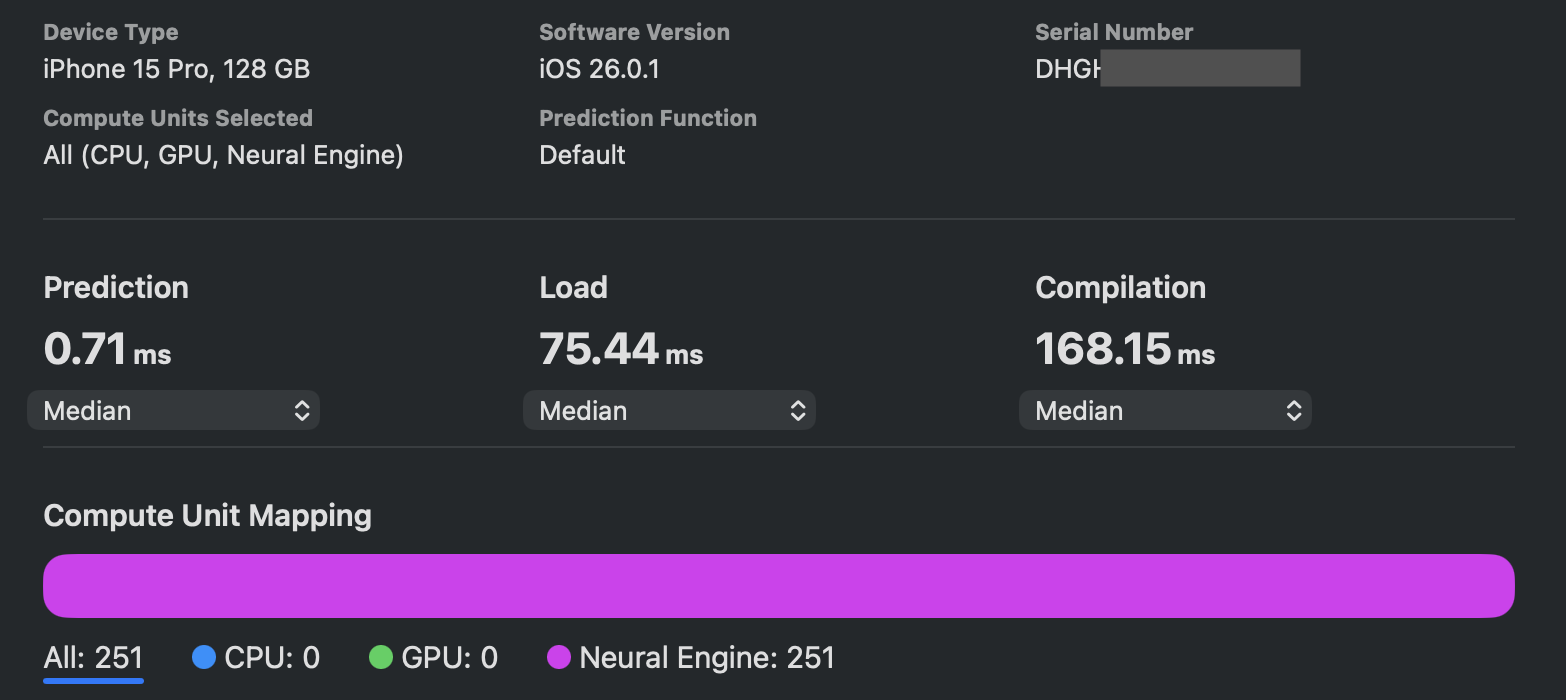}
    \\(4)\hspace{8cm} \\
    \caption{Latency test of (1) FaceLiVTv2-XS, (2) FaceLiVTv2-S, (3) FaceLiVTv2-M, (4). FaceLiVTv2-L with XCode using iPhone 15 pro. All model is cverted from pytorch framework to coremltols.}
    \label{fig:latency_test}
\end{figure}

\begin{table*}[!t]
\centering \footnotesize
\caption{Operations Category by Runtime Type}
\renewcommand{\arraystretch}{1.0}
\setlength{\tabcolsep}{6pt}
\begin{tabular}{|l|p{5cm}|p{10cm}|}
\hline
\textbf{Category} & \textbf{Operation Type} & \textbf{Description and Runtime Characteristic} \\ \hline

\multicolumn{3}{|l|}{\textbf{Compute Operations}} \\ \hline
Tensor MatMul & Linear projections and matrix multiplications & 
Dominated by arithmetic FLOPs and executed on Tensor Cores or ALUs. These operations are highly compute-bound and critical in attention and fully connected layers. \\ \hline

convolution & fully, depthwise and pointwise convolutions &
Performs spatial feature extraction using high multiply–accumulate (MAC) operations. Generally compute-intensive but cache-friendly and efficient on GPUs or NPUs. \\ \hline

Pooling & Global or average pooling operations &
Aggregates spatial information into compact representations. Has moderate compute cost with minimal memory overhead. \\ \hline

Activation  & Non-linear activation or attention weighting (GELU, ReLU, Softmax) &
Introduces non-linearity or attention weighting in neural modules. Arithmetic-dominated and often fused into preceding compute kernels for efficiency. \\ \hline

\multicolumn{3}{|l|}{\textbf{Memory Operations}} \\ \hline
Elementwise & Per-tensor arithmetic and residual connections (Add, Multiply, Broadcast) &
Lightweight arithmetic with frequent tensor reads and writes. These operations have low arithmetic intensity and are typically memory bandwidth limited. \\ \hline

Normalization & Mean–variance normalization across features (BN, LN, MVN) &
Applies scaling and centering to feature maps. Requires global memory access and synchronization, resulting in latency dominated by memory operations. \\ \hline

Copy/Reshape & Tensor layout changes and data movement (Reshape, Transpose, Concat, Split)&
Reorders, copies, or merges tensor data without significant computation. Memory-limited and responsible for large data transfer overheads, increasing runtime latency. \\ \hline
\end{tabular}
\end{table*}

All tests were conducted under identical conditions, with the device thermally stabilized and isolated from background processes to ensure consistent measurements. Each result represents the median value from 100 inference runs recorded by Xcode Instruments. As summarized in Fig.~\ref{fig:latency_test}, the FaceLiVTv2 models achieved prediction latencies of 0.43~ms, 0.54~ms, 0.65~ms, and 0.71~ms for the XS, S, M, and L variants, respectively. These results demonstrate that \textit{FaceLiVTv2} maintains excellent runtime scalability across model sizes while achieving sub-millisecond inference performance on Apple’s Neural Engine. 

\begin{figure}[!t]
    \centering
    \includegraphics[width=\columnwidth]{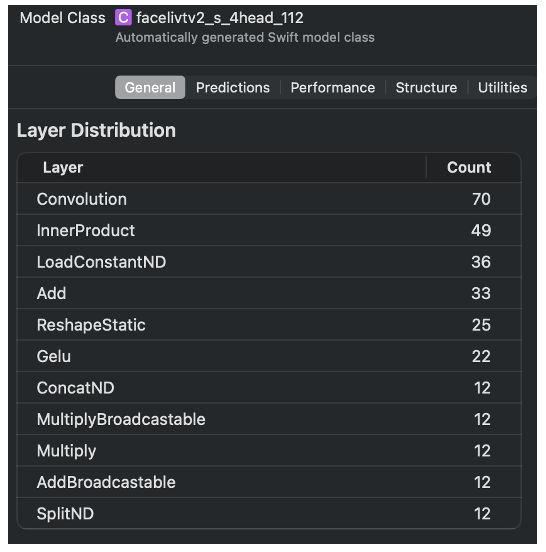}
    \caption{Layer distribution provided by XCode.}
    \label{fig:runtime_breakdown}
\end{figure}

\section*{Runtime Operation Breakdown Analysis}
To analyze the computational characteristics of each model, a runtime operation breakdown was conducted based on the Core~ML model profiling data. Each FaceLiVTv2 and baseline model (TransF-S, KANF-0.5, GhostFv1-1, and FaceLiVTv2-S) was first converted from the PyTorch framework into Core~ML format using \texttt{coremltools}. The resulting \texttt{.mlmodel} models were then profiled using \texttt{Xcode~Instruments} on an iPhone~15~Pro (A17~Pro chip) running \texttt{iOS~26.0.1}. During profiling, all inference operations were executed exclusively on the Neural~Engine, and the per-layer statistics were automatically logged by the Core~ML runtime analyzer. 

The \texttt{Layer Distribution} panel in Xcode was used to extract the frequency of each operator type, such as \texttt{convolution}, \texttt{InnerProduct}, \texttt{Add}, \texttt{GELU}, and \texttt{ReshapeStatic}. These counts were exported into CSV format and grouped into high-level categories of computational cost, including matrix multiplication (\texttt{Tensor~MatMul}), convolution, pooling, activation, and memory-related operations (elementwise, normalization, and copy/reshape). Each operation count was normalized to obtain its relative contribution percentage with respect to the total number of layers in the model. 

We visualize the runtime composition in Fig.~\ref{fig:breakdown} using the aggregated percentages of operation types from Fig.~\ref{fig:runtime_breakdown}. Compute-bound operations (\texttt{Tensor~MatMul}, \texttt{convolution}, \texttt{Pooling}, and \texttt{Activation}) are shown in blue and orange tones, while memory-bound operations (\texttt{Elementwise}, \texttt{Normalization}, \texttt{Copy/Reshape}) are displayed in lighter colors for contrast. The final figure summarizes the proportion of compute versus memory operations for each model, allowing direct comparison of architectural efficiency across Transformer-based and convolutional designs. 

\section*{ONNX Latency Evaluation Setup}
The inference latency of all FaceLiVTv2 variants was measured using RTX-5090 GPU, Intel i5-12500 with 64 GB of RAM, Jetson AGX Orin running with ONNX runtime without any quantization (FP32). Each model was first trained in the PyTorch framework and exported to ONNX using the \texttt{pytorch onnx} library. The following Python command was used to perform the conversion:
\begin{lstlisting}[language=Python, frame=None] 
model.to('cpu')
model = model.eval()
            
# batch size 1 for latency test
inp = torch.randn(1, 3, res, res, device='cpu')
torch.onnx.export(model, inp, f"./{name}_{1}.onnx", verbose = False, opset_version=12)

print("Finish...")
\end{lstlisting}

The converted ONNX models were deployed on-device with Linux based OS. Each model received a normalized 112$\times$112 RGB input tensor based on their specific input resolution, and the inference latency were measured using the ONNX runtime with CUDA core as the executor for GPU and Jetson AGX Orin GPU/CPU (MAX Power Mode = 50 Watt) with repetition 1000 image using this code:
\begin{lstlisting}[language=Python, frame=None] 
inp = np.random.rand(1, 3, 
      res, res).astype('float32')
timing = []
# warm-up
start = time.perf_counter()
while time.perf_counter() - start < T0:
    ort_inp = {model_ort.get_inputs()[0].name: inp}
    _ = model_ort.run([], ort_inp)

for i in range(repetition):
    start = time.perf_counter()
    ort_inp = {model_ort.get_inputs()[0].name: inp}
    _ = model_ort.run([], ort_inp)
    timing.append((time.perf_counter()-start)*1000.0)# ms

total_time_s  = np.sum(timing)/1000.0  # s
mean_lat_ms   = float(np.mean(timing))
std_lat_ms    = float(np.std(timing))

return {"model": name, "mean_lat": mean_lat,  
    "std_lat_ms": std_lat,  "total_time": total_time}
\end{lstlisting}

\begin{table*}[!t]
\centering
\setlength{\tabcolsep}{2.5pt}
\caption{Performance gap between lightweight FaceLiVTv2 variants and 
large-scale SOTA FR models. Mean Acc denotes the mean accuracy over 
LFW, CFP-FP, and AgeDB-30. FLOPs and mobile latency are measured 
at $112\times112$ input on iPhone 15 Pro.}
\begin{tabular}{lcccccccccccc}
\hline
\multirow{2}{*}{Model} & \multirow{2}{*}{Year} & Param & FLOPs & Training & \multirow{2}{*}{LFW$\uparrow$} & CFP & Age & Mean & \multicolumn{2}{c}{IJB (FAR@$10^{-4}$)$\uparrow$} & Latency \\ \cline{10-11}
& & (M) & (M) & Dataset & & -FP$\uparrow$ & DB30$\uparrow$ & Acc(\%)$\uparrow$ & B & C & ($ms$)$\downarrow$ \\
\hline
\multicolumn{12}{l}{\textit{Large-Scale FR Models (FLOPs $\geq$ 1G)}} \\
\hline
\rowcolor{gray!10}
ResNet200-TopoFR \cite{dan2024topofr}   & '24 & 118.8 & 23.5G & Glint360K & 99.87 & 99.45 & 98.82 & 99.38 & 97.84 & 96.56 & 12.24 \\
TransFace-L \cite{dan2025transface++}   & '25 & 271.6 & 25.4G & Glint360K & 99.85 & 99.37 & 98.66 & 99.29 & 96.73 & 97.85 & OOM   \\
TransFace-B \cite{dan2025transface++}   & '25 & 124.5 & 11.5G & Glint360K & 99.85 & 99.24 & 98.62 & 99.24 & 96.46 & 97.59 & 18.20 \\
TransFace-S \cite{dan2023transface}     & '23 & 86.7  & 5.8G  & Glint360K & 99.85 & 98.91 & 98.50 & 99.09 & 96.05 & 97.33 & 14.31 \\
ResNet50-ArcFace \cite{deng2019arcface} & '22 & 43.6  & 6.3G  & Glint360K & 99.78 & 98.77 & 98.28 & 98.94 & 95.30 & 96.81 & 3.76  \\
\hline
\multicolumn{12}{l}{\textit{Lightweight FR Models}} \\
\hline
\rowcolor{gray!10}
FaceLiVTv2-L  & - & 8.52 & 309 & Glint360K & 99.80\textbf{(-0.07)} & 98.26\textbf{(-1.19)} & 98.02\textbf{(-0.80)} & 98.69\textbf{(-0.69)} & 95.18\textbf{(-2.66)} & 96.59\textbf{(+0.03)} & 0.71\textbf{(17.2$\times\downarrow$)} \\
\rowcolor{gray!10}
FaceLiVTv2-M  & - & 7.02 & 258 & Glint360K & 99.78\textbf{(-0.09)} & 97.93\textbf{(-1.52)} & 98.10\textbf{(-0.72)} & 98.60\textbf{(-0.78)} & 95.02\textbf{(-2.82)} & 96.42\textbf{(-0.14)} & 0.65(18.8$\times\downarrow$) \\
\rowcolor{gray!10}
FaceLiVTv2-S  & - & 4.62 & 179 & Glint360K & 99.78\textbf{(-0.09)} & 97.47\textbf{(-1.98)} & 97.82\textbf{(-1.00)} & 98.36\textbf{(-1.02)} & 94.51\textbf{(-3.33)} & 95.99\textbf{(-0.57)} & 0.54(22.7$\times\downarrow$) \\
\rowcolor{gray!10}
FaceLiVTv2-XS & - & 2.9  & 90  & Glint360K & 99.63\textbf{(-0.24)} & 95.23\textbf{(-4.22)} & 96.68\textbf{(-2.14)} & 97.18\textbf{(-2.20)} & 90.67\textbf{(-7.17)} & 91.25\textbf{(-5.31)} & 0.43(28.5$\times\downarrow$) \\
\hline

\end{tabular}
\label{tab:gap_analysis}
\end{table*}

All tests were conducted under identical conditions, with the device thermally stabilized and isolated from background processes to ensure consistent measurements. Each result represents the mean value from inference runs for the latency test and a warm-up of 10 seconds (T0=10) for stability. As summarized in Table V of the main paper, the FaceLiVTv2 models achieved lower latency than the SOTA. These results demonstrate that \textit{FaceLiVTv2} maintains excellent runtime while achieving accuracy-latency efficiency across all devices. 

\section*{Performance Gap Analysis: Lightweight vs. Large-Scale Models}
To provide broader context for evaluating FaceLiVTv2, we compare its performance against large-scale SOTA face recognition models that are impractical for mobile and edge deployment due to their substantial computational demands. As summarized in Table~\ref{tab:gap_analysis}, we include five large-scale models — ResNet200-TopoFR\cite{dan2024topofr}, TransFace-L\cite{dan2025transface++}, TransFace-B\cite{dan2025transface++}, TransFace-S\cite{dan2025transface++}, and ResNet50-ArcFace\cite{deng2019arcface} — alongside the FaceLiVTv2 variants.

\section*{Accuracy Gap}

FaceLiVTv2-L demonstrates the most compelling accuracy-efficiency trade-off among the evaluated lightweight models. Against ResNet200-TopoFR, FaceLiVTv2-L narrows the mean accuracy gap to only $\mathbf{0.69\%}$ while requiring $\mathbf{13.9\times}$ fewer parameters and $\mathbf{76.1\times}$ fewer FLOPs. Remarkably,  FaceLiVTv2-L achieves 96.59\% on IJB-C, marginally \textbf{surpassing}  ResNet200-TopoFR by $\mathbf{+0.03\%}$, demonstrating competitive open-set verification capability despite its lightweight design. Against TransFace-L as the strongest Transformer-based model, FaceLiVTv2-L achieves a mean accuracy gap of only $\mathbf{0.60\%}$, while using $\mathbf{31.9\times}$ fewer parameters and $\mathbf{82.2\times}$ fewer FLOPs. On IJB-C, FaceLiVTv2-L trails TransFace-L by only $1.26\%$, further confirming its strong open-set recognition capability at a fraction of the computational cost. These results collectively demonstrate that FaceLiVTv2-L effectively bridges the performance gap between lightweight and large-scale face recognition architectures.

\section*{Latency and Deployability Gap}
Beyond accuracy, the deployability gap between large-scale and lightweight models is equally critical. ResNet200-TopoFR requires 12.24 ms mobile inference latency, TransFace-S requires 14.31 ms, and TransFace-B requires 18.20 ms on iPhone 15 Pro, while TransFace-L exceeds available memory entirely (OOM), rendering it completely infeasible for mobile deployment. In contrast, FaceLiVTv2-L achieves 0.71 ms mobile latency, a speedup of $\mathbf{17.2\times}$ over ResNet200-TopoFR, while FaceLiVTv2-M, FaceLiVTv2-S, and FaceLiVTv2-XS further reduce latency to 0.65 ms, 0.54 ms, and 0.43 ms, respectively, representing speedups of $18.8\times$, $22.7\times$, and $28.5\times$ over ResNet200-TopoFR. Compared to TransFace-L, which fails to run on mobile hardware due to memory constraints, all FaceLiVTv2 variants operate comfortably in real-time, highlighting the fundamental deployability advantage of the proposed architecture.

\end{document}